\documentclass[acmsmall,screen]{acmart}
\makeatletter
\def\runningfoot{\def\@runningfoot{}}
\def\firstfoot{\def\@firstfoot{}}
\makeatother
\renewcommand\footnotetextcopyrightpermission[1]{} 
\usepackage{fancyhdr}
\AtBeginDocument{%
  \providecommand\BibTeX{{%
    \normalfont B\kern-0.5em{\scshape i\kern-0.25em b}\kern-0.8em\TeX}}}

\usepackage{amssymb}

\usepackage{multirow}
\usepackage{xcolor}
\usepackage{colortbl}  
\usepackage{pdfrender}
\usepackage{bbding}
\usepackage{rotating}

\usepackage{bm}
\usepackage{amsmath}
\usepackage[caption=false,font=footnotesize,labelfont=sf,textfont=sf]{subfig}
\usepackage[switch]{lineno}
\usepackage{url}
\usepackage{graphicx}

\usepackage{booktabs}
\usepackage{arydshln}

\definecolor{ForestGreen}{RGB}{34,139,34}
\definecolor{myred}{RGB}{220,20,60}

\newcommand{\Mat}[1]{\mathbf{#1}}

\newcommand{\Set}[1]{\mathcal{#1}}

\definecolor{mygray-bg}{gray}{0.9}

\usepackage{pifont}

\newcommand{\ie}{\emph{i.e.}}
\newcommand{\eg}{\emph{e.g.}}
\newcommand{\etal}{\emph{et al.}}

\newcommand{\wrt}{\emph{w.r.t. }}
\newcommand{\cf}{\emph{cf. }}
\newcommand{\vs}{\emph{v.s. }}

\usepackage{adjustbox}

\usepackage{footmisc}

\makeatletter
\newcommand{\thickhline}{%
    \noalign {\ifnum 0=`}\fi \hrule height 1pt
    \futurelet \reserved@a \@xhline
}

\makeatother

\begin{document}

\title{Decomposed Prototype Learning for Few-Shot Scene Graph Generation}

\author{Xingchen Li}
\email{xingchenl@zju.edu.cn}
\affiliation{%
  \institution{Zhejiang University}
  \city{Hangzhou}
  \state{Zhejiang}
  \country{China}
  \postcode{310027}
}

\author{Jun Xiao}
\email{junx@zju.edu.cn}
\affiliation{%
 \institution{Zhejiang University}
  \city{Hangzhou}
  \state{Zhejiang}
  \country{China}
  \postcode{310027}
}

\author{Guikun Chen}
\email{guikun.chen@zju.edu.cn}
\affiliation{%
  \institution{Zhejiang University}
  \city{Hangzhou}
  \state{Zhejiang}
  \country{China}
  \postcode{310027}
}

\author{Yinfu Feng}
\email{fyf200502@gmail.com}
\affiliation{%
 \institution{Alibaba Group}
  \city{Hangzhou}
  \state{Zhejiang}
  \country{China}
  \postcode{310023}
}

\author{Yi Yang}
\email{yangyics@zju.edu.cn}
\affiliation{%
 \institution{Zhejiang University}
  \city{Hangzhou}
  \state{Zhejiang}
  \country{China}
  \postcode{310027}
}

\author{An-An Liu}
\email{anan0422@gmail.com}
\affiliation{%
 \institution{Tianjin University}
  \city{Tianjin}
  \country{China}
  \postcode{300072}
}

\author{Long Chen}
\authornote{Long Chen is the corresponding author.}
\email{zjuchenlong@gmail.com}
\affiliation{%
  \institution{Hong Kong University of Science and Technology}
  \city{Hong Kong}
  \country{China}
  \postcode{999077}
}

\renewcommand{\shortauthors}{Li and Xiao, et al.}

\begin{abstract}
Today's scene graph generation (SGG) models typically require abundant manual annotations to learn new predicate types. Therefore, it is difficult to apply them to real-world applications with massive uncommon predicate categories whose annotations are hard to collect.
In this paper, we focus on \emph{Few-Shot SGG (FSSGG)}, which encourages SGG models to be able to quickly transfer previous knowledge and recognize unseen predicates well with only a few examples.
However, current methods for FSSGG are hindered by the high intra-class variance of predicate categories in SGG:
On one hand, each predicate category commonly has multiple semantic meanings under different contexts. 
On the other hand, the visual appearance of relation triplets with the same predicate differs greatly under different subject-object compositions.
Such great variance of inputs makes it hard to learn generalizable representation for each predicate category with current few-shot learning (FSL) methods. However, we found that this intra-class variance of predicates is highly related to the composed subjects and objects.
To model the intra-class variance of predicates with subject-object context, we propose a novel \emph{Decomposed Prototype Learning (DPL)} model for FSSGG. Specifically, we first construct a decomposable prototype space to capture diverse semantics and visual patterns of subjects and objects for predicates by decomposing them into multiple prototypes.
Afterwards, we integrate these prototypes with different weights to generate query-adaptive predicate representation with more reliable semantics for each query sample.
We conduct extensive experiments and compare with various baseline methods to show the effectiveness of our method.
\end{abstract}



\maketitle

\section{Introduction}
Scene Graph Generation (SGG) is a challenging scene understanding task that aims to detect visual relationships between object pairs in one image and represent them in the form of a triplet: $\left \langle \texttt{sub}, \texttt{pred}, \texttt{obj} \right \rangle $. Recently, increasing works have been committed to this task and achieved great success, and most of them typically rely on abundant manual annotations.
However, the distribution of visual relationships in real-world scenarios is uneven, which involves vast amounts of less-frequently occurring predicates (or relationships) whose annotations are hard to collect.
Furthermore, due to the expensive annotation form ($N^{2}$ annotation cost)~\cite{li2022integrating} and the notorious multi-label noisy annotation issue~\cite{li2022devil}, labeling a high-quality SGG dataset for these rare predicates is much more difficult. As a result, most existing SGG models cannot achieve ideal performance on real-world applications with limited labels.

\begin{figure}[t]
    \centering
    \includegraphics[width=0.85\linewidth]{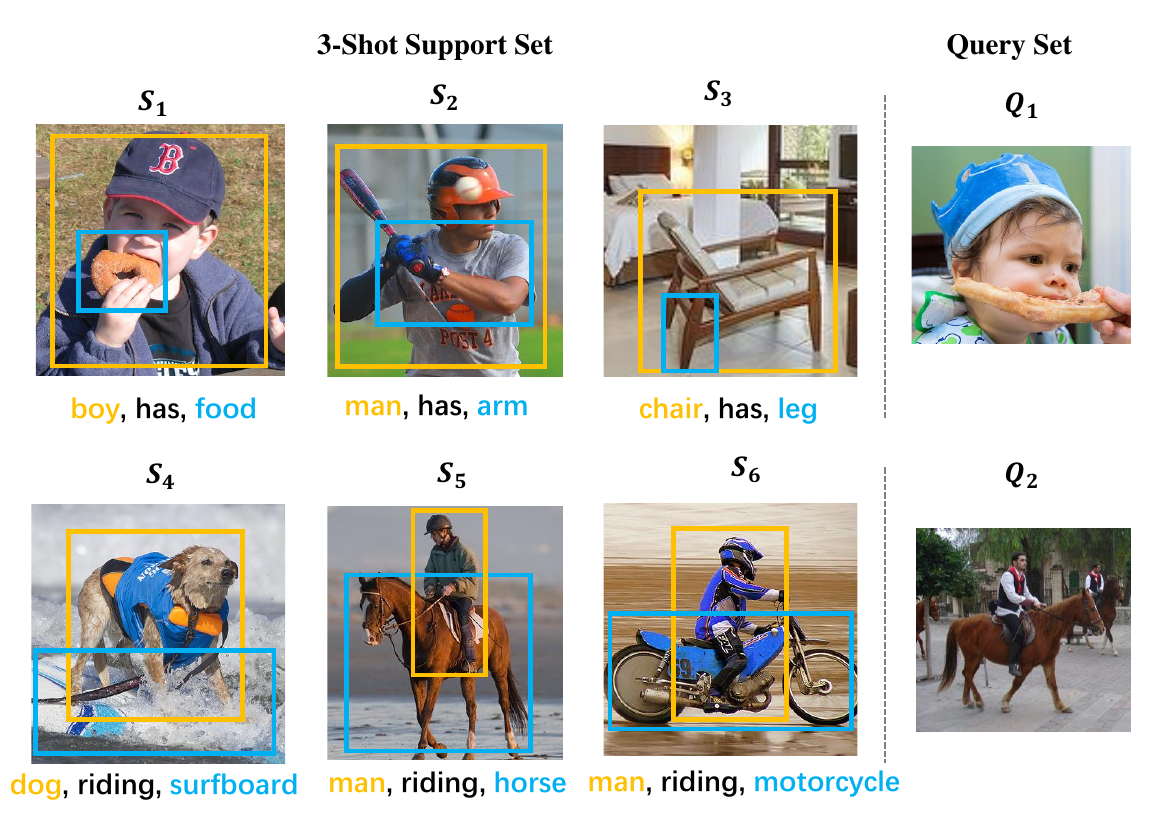}
    \caption{The illustration of FSSGG with 3-shots. For each predicate category, 3 support samples are provided with annotations (\ie, the bounding boxes, categories of subjects and objects, and their predicate categories). The target of FSSGG models is to detect the relation triplets in the query images.}
    \label{fig:intro}
\end{figure}

In contrast, we humans are able to generalize the previous knowledge on unseen visual relationships quickly with only a few examples and recognize them well.
In this paper, we aim to imitate this human ability and equip SGG models with this powerful transfer ability. Inspired by few-shot object detection~\cite{snell2017prototypical,sung2018learning} and segmentation~\cite{wang2019panet,li2021adaptive,zheng2022quaternion}, some recent studies~\cite{li2022zero,dornadula2019visual} attempt to improve SGG model performance in the few-shot setting, \ie, Few-Shot Scene Graph Generation (FSSGG).
As shown in Fig.~\ref{fig:intro}, referring to a few annotated examples (\ie, \textbf{support samples}) of an unseen (novel) predicate category, an ideal FSSGG model is expected to quickly capture the crucial clue for this predicate, recognize the relation triplets of it, and localize them in the target image (\ie, \textbf{query image}).  
Some efforts have been made to transfer the knowledge from seen to unseen predicates. For example, Dornadula \etal~\cite{dornadula2019visual} formulated each predicate as message passing functions between subjects and objects, Li \etal~\cite{li2022zero} modelled unseen predicates based on seen predicates by a memory mechanism.

While the main challenge of transferring knowledge to unseen predicates is the great intra-class variation of each predicate category.
On one hand, polysemy is very common in the predicate categories, and each predicate category may has diverse semantics under different contexts. For example, the predicate \texttt{has} may not only mean \emph{possession relation} between main body and individuals (\eg, $\left \langle  \texttt{chair}, \texttt{has}, \texttt{leg} \right \rangle $) but also mean \emph{eating relation} between human and food (\eg, $\left \langle  \texttt{boy}, \texttt{has}, \texttt{food} \right \rangle $) as shown in Fig.~\ref{fig:intro}. On the other hand, due to the compositional nature, the relation triplets of the same predicate sometimes have a totally distinct visual appearance under different subject-object pairs. For instance, the relation triplets of $\left \langle \texttt{man}, \texttt{has}, \texttt{arm} \right \rangle $ in $S_2$ and $\left \langle  \texttt{chair}, \texttt{has}, \texttt{leg} \right \rangle $ in $S_3$ look irrelevant visually. Such high variance of semantics and visual appearance for predicates makes it hard for current methods to discriminate representations for each predicate with limited labels and fail to generalize on novel subject-object pair compositions.

\begin{figure}[t]
    \centering
    \includegraphics[width=0.85\linewidth]{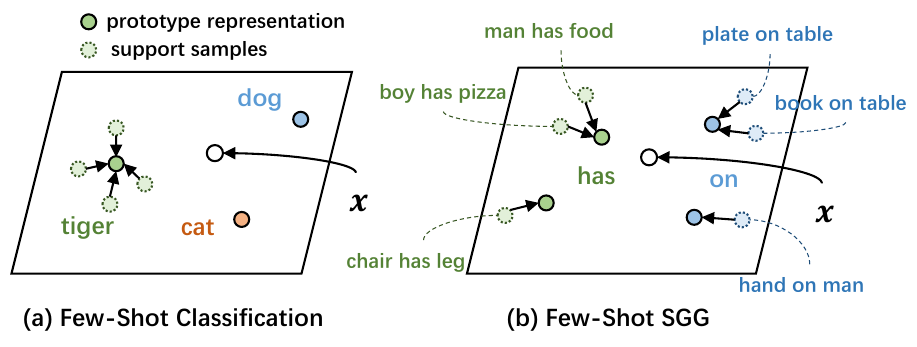}
    \caption{(a) The prototype representation for each class in conventional few-shot image classification. (b) Due to the high intra-class variance in FSSGG, the predicate may have multiple prototypes in the latent space.}
    \label{fig:intro2}
\end{figure}

To solve such intra-class variation, previous researchers have devised several methods to help generate a robust category prototype representation for each predicate category. For instance, Li \etal~\cite{li2022zero} proposed to distill lexical knowledge of different objects and embed them into the predicate prototype representations. In our work, we turn to model each predicate with multiple representations instead of carefully learning a single representation.
Assuming that we project each input sample ($\boldsymbol{x}$) into the latent space as shown in Fig.~\ref{fig:intro2}. In conventional few-shot classification, 
the samples of the same category tend to cluster around a central point (\ie, the prototype) as they usually share similar visual features~\cite{kim2019variational,snell2017prototypical,sung2018learning}. 
For FSSGG, it is difficult to find a unified representation for a predicate category due to the great variance of input visual features.
Besides, simply adopting a single prototype representation will cause the loss of semantic diversity of predicates under different contexts.
Therefore, we think that it would be more reasonable to decompose one predicate category representation into multiple prototypes to model the intra-class variance of predicates.

In this paper, we propose a novel Decomposed Prototype Learning (\textbf{DPL}) model for FSSGG. In order to model diverse intrinsic composition patterns of predicates, we construct a decomposable latent space and represent each predicate as multiple decomposed prototypes in this space.  
As the semantics and visual appearance of each predicate are highly relevant to the subject-object pairs, 
we exploit the subject-object context of support samples to help us construct predicate prototype representations.
However, due to the small number of support samples, it is difficult to catch enough information to build high-quality prototype representations.
Inspired by recent prompt-tuning works, we devise a set of learnable prompts to help us infuse the knowledge of pretrained Vision-Language (VL) models which have powerful expression ability of both vision and semantics.
Besides, in order to explore more novel compositions of object pairs for each predicate, we generate the subject prototype and object prototype separately based on each support sample (\eg, \texttt{boy}-\texttt{has} and \texttt{has}-\texttt{food} are regarded as two prototypes that can be composed with other subject or object prototypes).
These prototypes can be regarded as composable units to construct different semantics for each predicate category.
However, directly averaging these prototypes for each predicate is not able to handle different contexts of query samples.
Hence, we propose to generate query-adaptive prototype representations for each predicate with more appropriate semantics given different query samples. 
Specifically, for a pair of query sample and predicate category, we assign adaptive weights to different predicate prototypes according to their subjects and objects, and then exploit the weighted prototype representation as the final prototype representation to estimate the distance, thereby making more reliable predictions.

To better benchmark the Few-Shot SGG task, we re-split the prevalent Visual Genome (VG) dataset~\cite{krishna2017visual} into 80 seen predicate categories and 60 unseen predicate categories to evaluate the transfer ability of the model on novel predicate categories. Meanwhile, the object categories are shared in the samples of both seen and unseen predicate categories. Finally, we compare our proposed model with various FSL methods under different data splits and few-shot settings to show the effectiveness of our work.

We summarize our contributions as follows:
\begin{itemize}
    \item We focus on a promising research direction, FSSGG, and propose a novel Decomposed Prototype Learning (DPL) model, which can construct multiple prototype representations for each predicate category to represent its diverse semantics and different compositions of subjects and objects.
    \item We devise a set of compositional prompts to leverage the knowledge of the pretrained VL model to construct more powerful prototypes. Moreover, we generate query-adaptive predicate representations for each query sample by assigning different weights to each prototype according to the subject and object context.
    \item We conduct our experiments based on different datasets and few-shot settings. Extensive experimental results have shown the great superiority of our method.
\end{itemize}

\section{Related Work}

\subsection{Few-Shot Learning (FSL)}
FSL has been widely explored in various scene understanding tasks to get rid of costly manual labeling efforts.
It aims to learn valid knowledge about novel categories quickly with only a few reference samples. 
To achieve this goal, lots of creative solutions have been proposed.
One of the most well-known paradigms of FSL is meta-learning, which can be further categorized into three groups: 1) \emph{Metric-based}: Learning a projection function to map samples to the same embedding space and then measuring the similarity between query and support samples on the learned space~\cite{snell2017prototypical,sung2018learning,li2019distribution,li2019revisiting,jiang2020multi}. 
For example, Snell \etal~\cite{snell2017prototypical} assume that each category has one prototype representation in the embedding space. They obtained the prototype for each category by averaging the embedding representations of all support samples, and then measured the Euclidean distance between the query sample and the averaged prototype embedding to make prediction. 
Subsequently, Sung \etal~\cite{sung2018learning} proposed to estimate distance with a fully connected layer network instead of Euclidean distance calculation.
2) \emph{Model-based}: Designing model architectures to achieve quick adaptation on novel tasks. For example, Santoro \etal~\cite{santoro2016meta} proposed a memory-enhanced neural network to improve the model transfer ability on novel datas.
3) \emph{Optimization-based}: Learning a good model initialization that enables the model to be adapted to novel tasks using a few optimization steps~\cite{lee2019meta,finn2017model}.
Recently, some works have focused on prompt learning methods that design proper prompts for few-shot or zero-shot learning using large pretrained models~\cite {tsimpoukelli2021multimodal,alayrac2022flamingo,he2022towards,gu2021ppt,zhuounified}.

\subsection{Scene Graph Generation (SGG)}
SGG aims to predict visual relationship triplets between object pairs in an image. 
Recent SGG works mainly focus on three directions: 1) \emph{Model Architectures}: Current mainstream SGG methods are mostly based on two-stage: they first detect all localized objects and subsequently predict their entity categories and pair-wise relationships. Usually, they focus on modeling image context information~\cite{zellers2018neural,tang2019learning,yang2018graph,chen2019counterfactual,zheng2023prototype,zhang2023boosting}, language priors~\cite{lu2016visual,zellers2018neural} and so on.
Recently, several one-stage SGG works~\cite{li2022sgtr,teng2022structured} based on Transformer~\cite{vaswani2017attention} have been proposed and achieved impressive performance. They generate the pairs of localized objects and their relationships simultaneously. 2) \emph{Unbiased SGG}: Due to the long-tailed predicate distribution in the prevalent SGG datasets, unbiased SGG which aims to improve the performance of tail informative predicates has attracted considerable attention~\cite{tang2020unbiased,li2022devil,lyu2023adaptive,lyu2023generalized,zheng2023dual,li2023compositional,shi2024easy,li2024nicest,li2023label,li2022rethinking,chen2023addressing}. 3) \emph{Weakly-supervised SGG}: To overcome the limitation of expensive annotations, researchers proposed to solve SGG under a weakly-supervised setting, \ie, using image-level annotations as supervision~\cite{zhong2021learning,li2022integrating}. 

\noindent \textbf{Few-shot SGG.}
Few-shot SGG is first proposed by Dornadula \etal~\cite{dornadula2019visual} to improve the performance of those less-frequently occurring predicates in the world and in training data. To disentangle object representations from encoding specific information about a predicate, they formulate predicates as functions transform between objects and subjects.
Afterwards, some work followed this direction and did some research, however, the existing relevant research is still limited. Wang \etal~\cite{wang2020one} established two graph networks for the predicate category and the object category respectively to fuse contextual features between categories by information transformation. 
Yu \etal~\cite{yu2023knowledge} utilized both textual knowledge and visual relational knowledge to improve the generalization ability on unseen predicates. 
Besides, there are also some work focusing on zero-shot setting~\cite{li2023zero,li2022zero,chen2024learning}.
For example, to alleviate the difficulty of transferring knowledge from seen predicates to unseen predicates, Li \etal~\cite{li2022zero} distilled lexical knowledge of different objects and constructed the unseen predicates based on the seen predicates with the memory mechanism.
As the target of SGG is to detect triplet relationships with predicates, there is another few-shot setting of SGG, which regards each unique relationship triplet as one category~\cite{guo2020one}. In this setting, each predicate category may have multiple support samples in the one-shot setting as they have different relationship triplets.
In this work, we focus on the first setting to transfer knowledge to unseen predicates by devising effective prompts to extract knowledge from VL models.

\subsection{Prompt Learning}
Prompt learning is an emerging technique to allow large pretrained models to be able to perform few-shot or even zero-shot learning on new scenarios by defining a new prompting function.
With the emergence of large VL models like CLIP~\cite{radford2021learning}, prompt learning has been widely used in various tasks in the vision community, such as zero-shot image classification~\cite{radford2021learning}, object detection~\cite{du2022learning}, and captioning~\cite{chen2022visualgpt}.
Early works~\cite{nag2022zero} just exploit fixed prompts to exploit the knowledge of VL models, and researchers found that model performance is highly related to the input prompts. 
However, the engineering of designing hand-crafted prompts is extremely time-consuming.
To make it more efficient, recent works~\cite{zhou2022learning} attempt to model the prompts with a set of learnable vectors in continuous space and achieved great success.
Currently, many excellent works~\cite{li2021prefix, zhou2022learning,zhou2022conditional,gao2023compositional,ma2023understanding} based on task-specific prompt learning have been proposed.
In this work, we devise learnable prompts to help us generate semantic prototypes (for both subject and object) of predicates.

\section{Method}

\subsection{Problem Formulation}
Given an image, SGG aims to not only detect the objects in the target image but also predict the predicate between them.
In this paper, we focus on the few-shot learning of SGG. 
Given a set of object category $\Set{C}$ and a set of predicate category $\Set{R}$, the predicate categories are split into the unseen set $\Set{R}^{U}$ and the seen set $\Set{R}^{N}$ (\ie, $\Set{R} = \Set{R}^{N} \cup \Set{R}^{U}$). The training set only includes samples of seen categories.
During the evaluation, each predicate category $r \in \Set{R}$ is provided with an annotated support example set $\Set{S}_r$ with $K$ shots (\ie, $K$ support images with only 1 annotated triplet sample for each predicate)\footnote{In the following, we use ``a support sample" (or ``a support triplet") to denote the annotated triplet in the support image of one predicate type.\label{footnote:sample}}. 
And for each support triplet $\left \langle s,r,o \right \rangle \in \Set{S}_r$, it is provided with category annotations of subject and object $ c_s, c_o \in \Set{C} $, as well as the bounding boxes $b_s, b_o$ of the object pairs.
Given a query image $I$ and the support set of one category $r$, FSSGG aims to detect all the target relationship triplets belonging to the support category $r$ in the query image.
Following the two-stage SGG pipeline, we first detect a set of proposals $ \Set{B}$ with a pretrained object detector,
For each query sample $q$ with a pair of proposal $b_s,b_o \in \Set{B}$, we encode them into a feature projection network and obtain the query embedding. 
And for each target predicate category $r$, we decompose it into multiple prototypes and adaptively aggregate them to obtain context-aware category representation.
Then we evaluate the embedding distance to perform classification.
To mimic the evaluation setting, the support set and query set are usually randomly sampled in one batch during the training process. We list some key notations in Table~\ref{tab:notations}.

\begin{table*}[]
    \centering
    \caption{Summary of key notation definitions.}
    \vspace{-1em}
    \begin{tabular}{c|l}
    \hline\thickhline
        Notation & Definition \\
        \hline \hline
       $\Set{R}$ & The set of all predicate categories. \\
       $\Set{C}$ & The set of all object categories. \\
        $\Set{R}^{U}$, $\Set{R}^{N}$ & The set of unseen (U) or seen (N) predicate categories. \\
        $\Set{S}_r$ & The set of support triplet samples belong to predicate category $r$. \\
        $\mathcal{V}(r)$ &  The token embedding of category word of $r$.\\
        $\Mat{T}^{s}$, $\Mat{T}^{o}$ & The learnable prompt tokens to construct subject (s) or object (o) prompts. \\
        $\Mat{P}^{s}_k$, $\Mat{P}^{o}_k$ & The $k$-th prompt to extract contextual knowledge for subjects (s) or objects (o). \\
        
    \hline\thickhline
    \end{tabular}
    
    \label{tab:notations}
\end{table*}

\subsection{Decomposed Prototype Learning Network}
\begin{figure*}
    \centering
    \includegraphics[width=1.0\linewidth]{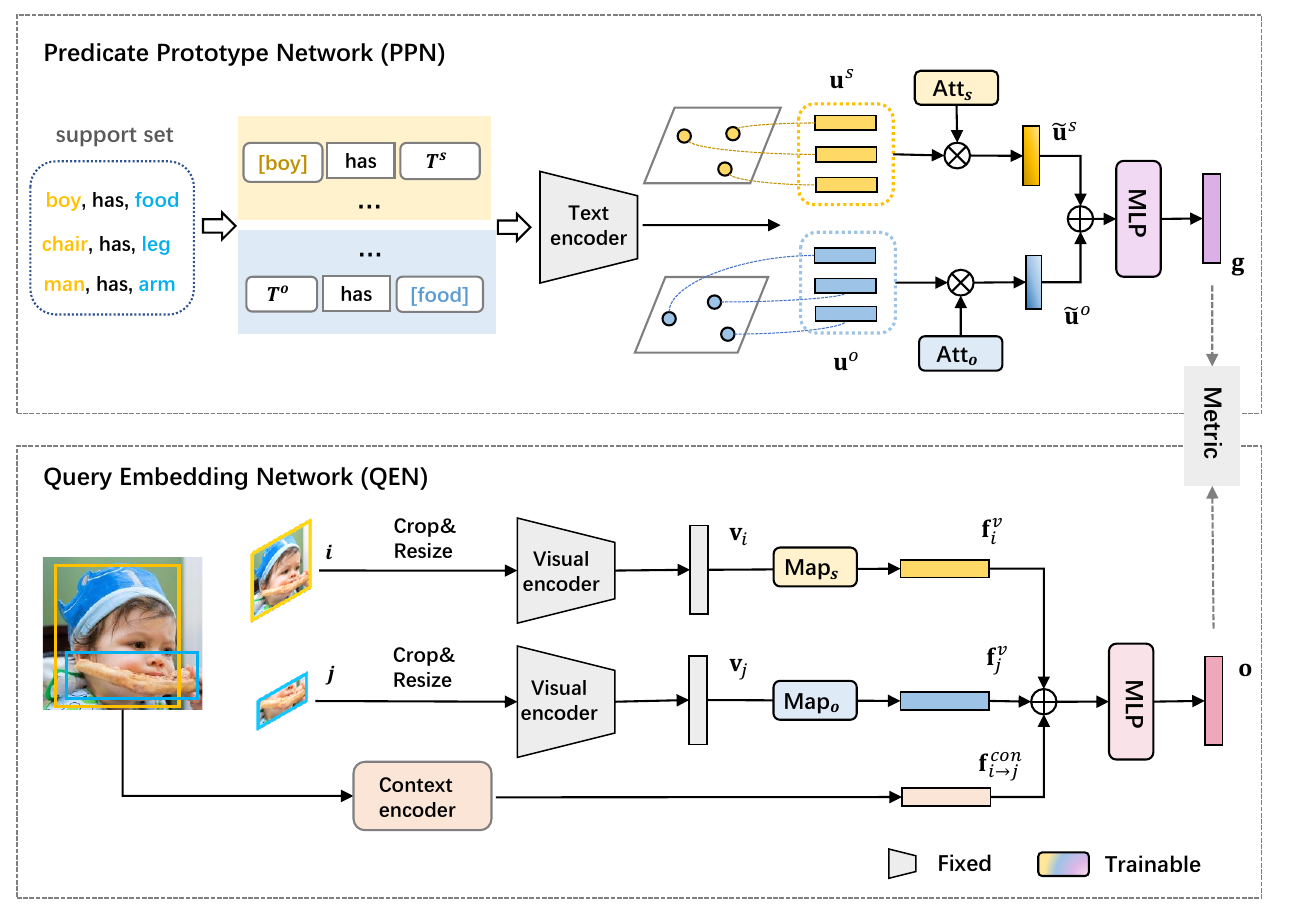}
    \caption{The overview of our Decomposed Prototype Learning Network. The whole framework consists of two main networks: 1) Query Embedding Network (QEN), which aims to generate embeddings of query samples, and 2) Predicate Prototype Network (PPN), which aims to generate query-adaptive predicate prototype representation of each target category for each input query sample. Then we estimate their distance with a metric learning method to perform classification. }
    \label{fig:framework}
\end{figure*}

Our model architecture mainly consists of two networks: 1) Query Embedding Network (QEN), which aims to project query samples to the same embedding space and generate their embeddings; 2) Predicate Prototype Network (PPN), which aims to generate multiple prototypes for each category based on their corresponding support samples, and then perform adaptive integration for each query sample to generate query-adaptive category representations. After obtaining the representations of the query sample and target category, we calculate the distance between them with a metric learning method to classify the query sample. The whole architecture is shown in Fig.~\ref{fig:framework}. In the following, we will introduce the two networks and the metric learning method respectively.

\subsubsection{\textbf{Query Embedding Network}} Given a query triplet sample $q$ in an image $I$, assuming that the proposals of its subject $i$ and object $j$ are $b_i$ and $b_j$ respectively, we first learn an embedding network to project the features of the query sample $q$ into the same latent space of the predicate representation, thereby evaluating its distance from the generated predicate category representation for classification.

\paragraph{Visual Encoder.}
In order to take advantage of the knowledge from the pretrained Vision-Language (VL) model, we first crop the subject $i$ and object $j$ according to their bounding boxes and then send them into the visual encoder $\text{Enc}_{vis} $ of the VL model to obtain the initial visual features of the input object pair:
\begin{equation} 
    \Mat{v}_{i} = \text{Enc}_{vis} \left (b_i \right ), \quad \Mat{v}_{j} = \text{Enc}_{vis} \left (b_j \right ). 
\end{equation}

Then we send the extracted visual features of the input object pair into the subject and object aware encoders respectively (\ie, $\text{Map}_{s}$ and $\text{Map}_{o}$) to obtain corresponding representations which will participate in the subsequent embedding encoding process: 
\begin{equation} \label{eq:visual}
    \Mat{f}^{v}_{i} = \text{Map}_{s} \left (\Mat{v}_{i} \right ), \quad \Mat{f}^{v}_{j} = \text{Map}_{o} \left (\Mat{v}_{j} \right ) ,
\end{equation}
where $\text{Map}_{s}$ and $\text{Map}_{o}$ are both two-layer Multi-Layer Perceptrons (MLP).

\paragraph{Context Encoder.}
As the predicate prediction for the query subject and object pair is highly relevant to its image scene context information (\eg, spatial positions of objects and global interaction information among other detected objects in the image), we follow the previous SGG works and exploit a context encoder $\text{Enc}_{con}$ to extract contextual information in the image. 
Given an arbitrary context encoder $\text{Enc}_{con} $ (such as Bi-LSTM~\cite{zellers2018neural}), we can obtain the context feature of each pair of objects $i$ and $j$:
\begin{equation}
    \Mat{f}^{con}_{i\rightarrow j} = \text{Enc}_{con} \left ( b_i, b_j, \Set{B}_{/ij} \right ),
\end{equation}
where $ \Set{B}_{/ij} $ is the set of other detected objects in image $I$.

\paragraph{Feature Aggregation.}
After obtaining these encoded features, we integrate them with an aggregator $\text{AGG}$:
\begin{equation}\label{eq:agg}
    \Mat{o}_{q,i\rightarrow j} = \text{AGG} \left ( \left [ \Mat{f}^{v}_{i}; \Mat{f}^{v}_{j}; \Mat{f}^{con}_{i\rightarrow j} \right ] \right ).
\end{equation}
Here we concatenate the encoded features together and exploit a two-layer MLP as AGG aggregator to generate $\Mat{o}_{q,i\rightarrow j}$, which is the embedding of the query sample with the subject $i$ and the object $j$. For the sake of simplicity, we will later omit the subscript $\scriptsize{i \to j}$ representing the object-subject pair $i,j$ in the query sample $q$, which is abbreviated as $\Mat{o}_{q} $, and this abbreviation will be used in the following text.

\subsubsection{\textbf{Predicate Prototype Network}}\label{sec:prototype}
In this section, we will introduce the details of our Predicate Prototype Network (PPN) based on our Decomposed Prototype Learning (DPL) method. We construct a decomposable semantic space for predicates with the help of the pretrained VL model. For each predicate category, we generate multiple prototypes related to subjects and objects based on their support samples. Given a query sample, we adaptively integrate these prototypes with assigned weights based on subject and object pair to generate query-adaptive predicate prototype representation for each query sample.

\paragraph{Prompt Construction}
Prompt tuning is an efficient way to transfer the knowledge of the large pretrained VL models into downstream tasks. In this paper, we introduce learnable prompts as indicators to help us construct the semantic space for each predicate with the power of VL models.
As the semantics of one predicate is highly related to the subject-object context, we combine the subject and object category with the prompts to infuse the semantic knowledge of predicates. To explore more compositions of subject-object context, we exploit the subjects and objects to generate prototypes separately.

Specifically, we first define the contextual prompt tokens $\Mat{T}^{s}= [\Mat{t}^{s}_1, \cdots, \Mat{t}^{s}_{L} ] \in \mathbb{R}^{L\times D}$ and $\Mat{T}^{o}= [\Mat{t}^{o}_1, \cdots, \Mat{t}^{o}_{L} ]\in \mathbb{R}^{L\times D}$ for the subject and object separately, and $L$ is the length of prompt tokens. 
Given a predicate category $r$ and one of its support triplet $k$, assuming that the subject-object pair of the triplet is $\left \langle x, y \right \rangle$ and their annotated categories are $ C_x, C_y$, we combine the token embeddings of $ C_x, C_y$ with our prompt tokens to construct the prompts, which will be sent into the following VL model to extract text embeddings:
\begin{equation}\label{eq:prompt1}
    \Mat{P}^{s}_{k} = \left[\mathcal{V}(C_x); \mathcal{V}(r); \Mat{T}^{s} \right], \quad \Mat{P}^{o}_{k} = \left[\Mat{T}^{o}; \mathcal{V}(r); \mathcal{V}(C_y) \right],
\end{equation}
where $\mathcal{V}(\cdot)$ is the embedding dictionary of all the word tokens. $\mathcal{V}(C_x)$ and $\mathcal{V}(C_y)$ are corresponding token embeddings of category words for subject $x$ and object $y$ respectively. And $\mathcal{V} (r)$ is the word token embedding of the predicate category $r$. The above token embeddings are then combined together with the learnable prompt tokens $\Mat{T}^{s}$ and $\Mat{T}^{o}$ to construct the complete prompts, which will be sent into the text encoder of VL model.

However, directly utilizing the word embeddings of the subject and object categories (\ie, $\mathcal{V}(C_x)$ and $\mathcal{V}(C_y)$) to construct the prompts may cause the meaning of the extracted prototypes limited to one specific subject or object category. In fact, we expect that generated prototypes can contain the information about those semantically similar categories to the subject $x$ and object $y$, thereby receiving better generalization ability of the obtained prototype representations. Therefore, we introduce a set of learnable vectors $\Mat{H}=\left \{ \Mat{h}_{C_1},\dots,\Mat{h}_{|\Set{C}|} \right \}$ to learn the representation of each object category to capture their correlations. We build a fully connected graph to connect all object categories and the edges represent the similarity among them. The nodes of the graph represent different object categories, and the weights of their edges are calculated by the cosine similarity of node representations. Then, for each node $i$, we can update its representation $\Mat{h}_i$ based on its neighbors and edge weights:
\begin{equation}
    \Mat{h}'_i = \Mat{h}_i + \sum_{j\in \Set{N}_i}\tilde{w}_{ij}\Mat{h}_j, \quad \tilde{w}_{ij} = \text{Softmax}_{i}\left (\text{Cosine} \left (\Mat{h}_{i}, \Mat{h}_j \right ) \right )
\end{equation}
where the updated node representation $\Mat{h}'_i$ integrates its semantically similar category information, 
$\tilde{w}_{ij}$ represents the normalized weight scores between node $i$ and node $j$ for each predicate category after softmax, $\Set{N}_i$ is the set of neighbor nodes of $i$, excluding node $i$ itself. To endow learnable vectors with semantic information of objects, we initialize them with the semantic embedding vectors\footnote{To obtain the semantic embedding vectors of each object category, we use a predefined fixed prompt to input the predicted object categories of the subject and object into the VL model to obtain the text embedding representation. Here we exploit a fixed prompt ``\texttt{This is a photo of [cls]}" and \texttt{[cls]} is the word tokens of predicted object categories.\label{fn:sem-vec}} of the corresponding category.

In this way, the prompt representation in Eq.~\eqref{eq:prompt1} can be transformed into the following format:
\begin{equation}
    \Mat{P}^{s}_{k} = \left[\Mat{h}'_{C_x}; \mathcal{V}(r); \Mat{T}^{s} \right], \quad \Mat{P}^{o}_{k} = \left[\Mat{T}^{o}; \mathcal{V}(r); \Mat{h}'_{C_y} \right],
\end{equation}
where $\Mat{h}'_{C_x}$ and $\Mat{h}'_{C_y}$ respectively denote the corresponding category node representations of subject $x$ and object $y$ after updating.

\paragraph{Prototype Representation Generation}
After obtaining the prompts according to all support samples of each category, we input them into the text encoder $\text{Enc}_{txt}$ of the VL model to extract the text embedding of each prompt:
\begin{equation}
    \Mat{u}^{s}_k = \text{Enc}_{txt} \left ( \Mat{P}^{s}_k \right ), \quad \Mat{u}^{o}_k = \text{Enc}_{txt} \left ( \Mat{P}^{o}_k \right ),
\end{equation}
where $\Mat{P}^{s}_k \in \Set{P}^{s}$ and $\Mat{P}^{o}_k \in \Set{P}^{o}$ respectively denote the prompt representations generated from the subject and object of the support sample $k$.
The obtained text embeddings $\Mat{u}^{s}_k \in \Set{U}^{s}$ and $\Mat{u}^{o}_k \in \Set{U}^{o} $ can be regarded as the semantic knowledge extracted from the VL model, which are the prototype representation vectors of the predicate category.

Since the predicate triplet samples may have different semantics under different subject-object pairs, we adaptively assign different weights to each prototype to generate more reliable prototype representations for each query sample. Here, we mainly consider the semantic similarity of the subject and object between the query sample and the support sample, which can be calculated by the similarity of the text embedding vectors of the subject and object categories.
Assume that the annotated categories of subject and object of the support sample $k$ are denoted as $C^s_k$ and $C^o_k$, and the predicted categories of subject and object by query sample $q$ are denoted as $C^s_q$ and $C^o_q$ (the details of object category prediction is in Sec~\ref{sec:obj-cls}.).
After obtaining the categories of subject and object, we calculate the similarity scores of their semantic embedding vectors\footref{fn:sem-vec} respectively, formulated as:
\begin{equation}
    a_k^s = \text{Att}_{s}(C^s_q, C^s_k), \quad a_k^o = \text{Att}_{o}(C^o_q, C^o_k).
\end{equation}
Here we exploit the cosine similarity of the embedding vectors to calculate the similarity scores. For simplicity, we omit the subscript of the query sample $q$.
In order to avoid the dimensional impact of the data, we first use the Min-Max Normalization method to scale the similarity score, and then use the softmax function to normalize the score $a^{*}$ to obtain the attention scores $\tilde{a}^{*}$ (including the subject and object) of the query sample with all prototypes.

Afterwards, we aggregate the generated subject and object prototype representations $\Mat{u}$ with the normalized attention scores $\tilde{a}$:
\begin{equation}
    \tilde{\Mat{u}}^{s} =  { \sum_{k=1}^{|\Set{P}|}} \tilde{a}^{s}_{k} \Mat{u}^{s}_k, \quad \tilde{\Mat{u}}^{o} =  { \sum_{k=1}^{|\Set{P}|}} \tilde{a}^{o}_{k} \Mat{u}^{o}_{k},
\end{equation}
where $|\Set{P}|$ is the number of subject and object prototypes. Afterwards, we concatenate the weighted prototype representations of the subject and object together:
\begin{equation}
    \Mat{f}^{pro}=[\tilde{\Mat{u}}^{s}; \tilde{\Mat{u}}^{o}].
\end{equation} 
Subsequently, $\Mat{f}^{pro}$ is fed into a mapping network to obtain the query-adaptive predicate category representation for a query sample:
\begin{equation}
    \Mat{g}_r=\text{Gen}(\Mat{f}^{pro}),
\end{equation}
where $\text{Gen}$ is a two-layer Fully Connected (FC) layer. After obtaining the predicate prototype representation, we calculate its distance with the query sample embedding to perform classification. Here we omit the subscript representation of the query sample. Note that for the query sample $q$, the complete form of the generated query-adaptive prototype representation of predicate $r$ is $\Mat{g}_{q,r}$.

\subsubsection{\textbf{Metric learning}}
After obtaining the projected embedding $\Mat{o}_q$ of the query sample $q$ and its corresponding prototype representation $\Mat{g}_{q,r}$ of the predicate $r$, we calculate their cosine similarity scores as the metric learning method:
\begin{equation}
    d(\Mat{o}_q,\Mat{g}_{q,r}) = \tau \cdot \text{Cosine} \left ( \Mat{o}_q,\Mat{g}_{q,r} \right),
\end{equation}
where $\tau$ is the temperature parameter which controls the range of the logits in the following softmax~\cite{radford2021learning}.
For each query sample, we calculate its distance with all candidate predicate categories and select the closest category as the classification result.

\subsection{Model Training}
The overall training objective of our method consists of three parts: 1) the training objective of prompt learning for our decomposed prototype generation; 2) the objective of predicate category classification; and 3) the objective of object category classification. The total training loss function can be formulated as:
\begin{equation}
    \mathcal{L}=\mathcal{L}_{rel} + \mathcal{L}_{pro} + \mathcal{L}_{obj},
\end{equation}
where $\mathcal{L}_{pro}$, $\mathcal{L}_{rel}$ and $\mathcal{L}_{obj}$ are the training loss for prototype learning, predicate and object classification respectively. We introduce the details of them in the following.

\subsubsection{\textbf{Prompt Learning}}
In order to train the learnable prompt tokens for decomposed prototype generation, we utilize the support samples in the training phase, which are randomly sampled in each batch, to help us build the training objective. For one support sample $e$ sampled in a batch, we send it to the same embedding encoding network QEN as the query samples to obtain its encoded embedding $\Mat{o}_e$, and meanwhile generate its specific predicate prototype representations for all the candidate predicate categories in the batch.
Then we calculate the distance between the sample embedding and the category prototype representation. The goal of the training objective is to make the embedding of support sample $e$ close to its own generated category representation, and at the same time enlarge its distance from category representations generated by other support samples. To this end, we utilize the triplet loss function to calculate $\mathcal{L}_{pro}$:
\begin{equation}
    \mathcal{L}_{pro} = \sum_{e \in \Set{S}_b} \sum_{j \in \Set{S}_{b/e}} \alpha_{ej} \text{log}\left ( 1 + \text{exp} \left ( \delta_{e,j} \right ) \right ), 
\end{equation}
\begin{equation}
    \delta_{e,j}=d(\Mat{o}_e,\Mat{g}_{j,R_j})-d(\Mat{o}_e,\Mat{g}_{e,R_e}),
\end{equation}
where $\Set{S}_b$ is all the support samples in a batch $b$, $\Set{S}_{b/e}$ is the set of support samples in the batch excluding sample $e$ and is regarded as the set of negative samples of $e$, $\Mat{g}_{e,R_e}$ is the category representation of $R_e$ generated by the support sample $e$ itself, $R_e$ is the true predicate category label of the sample $e$, $\Mat{g}_{j,R_j}$ is the category representation of annotated label generated from any support sample $j$ in the batch.
And $\alpha_{ej}$ assigns different weights to each positive and negative sample pair. When $R_e$ and $R_j$ are the same category, $\alpha_{ej}$ is set smaller to give the sample pair a minor penalty, and when $R_e$ and $R_j$ are different categories, $\alpha_{ej}$ is set larger. We set $\alpha_{ej}$ as $0.5$ and $1$ in these two situations respectively in the experiment.

\subsubsection{\textbf{Predicate Classification}}
During the training process, we randomly sample the predicate categories from the seen categories $\Set{R}_N$ as the training set in a batch. For each predicate category $r$, we randomly select a part of its positive samples as the support set, and the remaining positive samples are used as query samples.
For each query sample $q$ in a batch, we calculate its distance to the positive category $r$ and the negative categories and apply cross-entropy loss:
\begin{equation}
    \mathcal{L}_{rel} = - \sum_{q \in \Set{D}_b} \left ( \text{log}\frac{\text{exp}( d(\Mat{o}_q,\Mat{g}_{q,r}))}{ {\textstyle \sum_{j \in \Set{R}_b}}\text{exp} ( d(\Mat{o}_q,\Mat{g}_{q,j})) } \right ),
\end{equation}
where $\Set{D}_b$ is all the query samples in a batch $b$, the embeddings $\Mat{g}_{q,j}$ and $\Mat{g}_{q,r}$ are the generated predicate representations of categories $j$ and $r$ respectively based on the query sample $q$.

\subsubsection{\textbf{Object Classification}}\label{sec:obj-cls}
For the classification of object categories, we use the same object context encoder and object classifier as Zellers \etal~\cite{zellers2018neural}.
We encode the context information of the entire image with a Bi-LSTM and express all the detected proposal information of the image in the form of a sequence: $\left [ (b_1,\Mat{v}_1,\Mat{l}_1),\dots, (b_n,\Mat{v}_n,\Mat{l}_n) \right ] $. Wherein $n$ is the number of all the object proposals detected in the image, and $\Mat{l}_i$ is the predicted label probability distribution of the detected proposal $i$ before context encoding. The object category probability are formulated as follows: 
\begin{equation}\label{eq:obj_cls}
    \Mat{Z} = \text{Cls}_{obj}(\text{Enc}_{obj}(\left [ (b_i, \Mat{v}_i, \Mat{l}_i) \right ]_{i=1,\dots ,n})),
\end{equation}
where $\Mat{Z}=\left [ \Mat{z}_1,\dots,\Mat{z}_n \right ] $ is the category probability distribution of each object output. More details about $\text{Enc}_{obj}$ and $\text{Cls}_{obj}$ can be found in~\cite{zellers2018neural}.
We take the category with the largest probability score as the final output object category.
We also calculate the loss function $\mathcal{L}_{obj}$ by using cross-entropy to train the object classification network.

\section{Experiments}

\subsection{Experimental Settings}
\subsubsection{\textbf{Datasets}} We conducted experiments with different settings on two most prevalent SGG datasets: Visual Genome (VG)~\cite{krishna2017visual} and GQA~\cite{dong2022stacked}:

\begin{enumerate}
    \item \textbf{VG-25:} To compare with existing methods, we conducted our experiment on the popular split with a total of 150 object categories and 50 predicate categories, including 25 most frequently seen predicates and 25 unseen predicates. The number of object categories is $150$.
    \item \textbf{VG-60:} Since the number of predicates in VG-25 is a bit small for a reliable evaluation on FSSGG models, we re-split the original VG dataset to get a larger dataset. We set $80$ common predicate categories as seen predicates and $60$ rare predicate categories as unseen predicates. We also selected the most common $247$ object categories.
    \item \textbf{GQA-50:} For GQA dataset, we follow the existing work~\cite{dong2022stacked} to utilize the split with $200$ object categories and 100 predicate categories. We set $50$ seen predicate categories and $50$ unseen predicate categories by their frequency.
    
\end{enumerate}

For all splits, the object categories in training and test sets are the same. The training set only includes the samples of seen predicate categories.
For each predicate category, we randomly split $K$-shot from its annotated samples as its support samples in the test set. In experiments, we set $K$ as $1$, $5$ and $10$, respectively.
The triplet samples annotated with the unseen predicate categories in the test set are called \emph{unseen set}, and the triplet samples annotated with the seen predicate categories are called \emph{seen set}.

\subsubsection{\textbf{Tasks}}
Following the previous work~\cite{li2022zero,dornadula2019visual}, we mainly evaluated FSSGG performance on the task \emph{Predicate Classification} (\textbf{PredCls}): Given the ground-truth object boxes and class labels, models are required to predict the predicate class between pairwise objects. 
For a complete comparison, we also evaluated the performance on task \emph{Scene Graph Classification} (\textbf{SGCls}): Given the ground-truth object boxes, models are required to predict object classes and predicate classes between each pair of objects.

\subsubsection{\textbf{Evaluation Metrics}}
Following prior studies~\cite{li2022zero,dornadula2019visual}, we reported model performance on Recall@N (R@N). To avoid the major influence of common categories on performance and evaluate each category equally, we also reported the performance on mean Recall@N (mR@N) which is the average of recall scores computed for each category separately. Following the previous FSL work~\cite{gidaris2018dynamic,kukleva2021generalized}, we report not only the performance of the model on unseen sets, but also the performance on seen sets to more comprehensively evaluate the model performance.

\subsubsection{\textbf{Training Details}}
To mimic the evaluation setting, we randomly generated the support set and query set from training data in a batch. For each batch, we selected $N_b$ categories and split its positive samples into the support set and query set randomly. 
The number of query and support samples for each predicate, and the number of sampled categories are set differently according to the batch data.
We set the ratio of foreground samples and background samples as 1:2. 
In experiments, we set the number of learnable tokens for each prompt as $24$ for all prompt-based methods.
We exploited the pretrained CLIP~\cite{radford2021learning} as our VL model in all our prompt-based experiments.
\subsubsection{\textbf{Baselines}}
To completely evaluate our method, we first compare with existing \emph{SGG methods in the few-shot setting}: \emph{\textbf{LPaF}}~\cite{dornadula2019visual}, which trains a two-layer MLP to predict the predicate category for the input object pair. \emph{\textbf{LKMN}}~\cite{li2022zero}, which introduces lexical knowledge to construct multi-modal representations of objects. 
\emph{\textbf{KFV}}~\cite{yu2023knowledge}, which takes advantage of VL models by constructing knowledge graphs and effective prompts. We also compare with a state-of-the-art SGG model \emph{\textbf{PENET}}~\cite{zheng2023prototype}, which represents each predicate by one prototype representation. We train it in the few-shot setting. We first trained the model with the training set and then fine-tuned the model with the support samples of unseen predicates in the test set.

We also compare with two popular \emph{metric learning methods}: 
\emph{\textbf{ProtoNet}}~\cite{snell2017prototypical}, which projects query samples and support samples into the same embedding space and then estimates the similarity between embedding vectors by calculating their Euclidean distance.
\emph{\textbf{RelNet}}~\cite{sung2018learning}, which also learns the embedding space for query and support samples, while introduces a trainable metric network to estimate the similarity between them.

Besides, we also compare with recent \emph{Vision-Language based methods}: \emph{\textbf{CoOp}}~\cite{zhou2022learning}, which introduces learnable prompts and utilizes the pretrained model CLIP to extract the text embedding of the target category, and then calculates its cosine similarity scores with the visual representation vector of each query sample.
\emph{\textbf{VinVL}}~\cite{zhang2021vinvl}, which was pretrained on millions of image-text pairs with fine-grained object-level annotations.
\emph{\textbf{TCP}}~\cite{yao2024tcp}, which incorporates class-level textual knowledge generate class-aware textual tokens and integrates these class-aware prompts into the text encoder.

Among these methods, \emph{LPaF}, \emph{LKMN} and \emph{CoOp} are pretrained on a training set containing only seen categories, and then fine-tuning strategy is performed on a small sample support set of unseen categories in the test set. \emph{ProtoNet} and \emph{RelNet} are only trained on the training set without fine-tuning on the support samples during evaluation.
In order to compare with these methods, we adopt two different training strategies. One is to train only on the training set containing samples of seen predicate categories, denoted as \textbf{\emph{DPL}}. The other is to perform pre-training on the training set and then perform the fine-tuning process on the few-shot support samples of unseen predicate categories, denoted as \textbf{\emph{DPL\textsuperscript{$\dagger$}}}. Note that only the objective function $\mathcal{L}_{pro}$ is optimized during fine-tuning process for \emph{DPL\textsuperscript{$\dagger$}}.
In order to make a fair comparison, for \emph{ProtoNet} and \emph{RelNet}, we adopted the same sampling strategy as \emph{DPL}.
In our experiments, we used the same Bi-LSTM model as our context encoder for all methods following the classic SGG model Motifs~\cite{zellers2018neural}. 

\begin{table}[]
    \centering
    \caption{Unseen predicate performance (\%) comparison on VG-25 dataset \wrt PredCls with K-shot. The \textcolor{red}{\textbf{best}} and \textcolor{blue}{second best} results are marked according to formats.}
    \vspace{-1em}
    \scalebox{1.0}{
    \begin{tabular}{ c |c |l l | c c c | c c c}
    \hline\thickhline
     \multirow{2}{*}{D} & \multirow{2}{*}{K-shot} & \multicolumn{2}{c|}{\multirow{2}{*}{Method}} & \multicolumn{3}{c|}{Recall} &  \multicolumn{3}{c}{meanRecall} \\
     \cline{5-10}
      & & & & @20 & @50 & @100  & @20 & @50 & @100  \\
  \hline\hline
   \multirow{24}{*}{\begin{sideways}VG-25\end{sideways}} & \multirow{7}{*}{1-shot} & 
     RelNet~\cite{sung2018learning} & $_{\textit{CVPR'18}}$  &  11.05 & 15.33 & 18.22 & 10.59 & 14.30 & 16.87 \\
   & & ProtoNet~\cite{snell2017prototypical} & $_{\textit{NIPS'17}}$  &  11.48 & 16.27 & 19.56 & 11.51 & 15.28 & 17.80 \\
    & & CoOp~\cite{zhou2022learning} & $_{\textit{IJCV'22}}$  &  17.52 & 21.86 & 24.17 & 16.55 & 20.64 & 22.45\\
    & & PENET~\cite{zheng2023prototype} & $_{\textit{CVPR'23}}$  & 21.29  & 26.40  & 28.78  & 23.02  &  27.39 & 29.45 \\
    & & TCP~\cite{yao2024tcp} & $_{\textit{CVPR'24}}$ & 18.46 & 23.30  & 25.68 &  18.28 & 22.69  & 24.82  \\
    & & \cellcolor{mygray-bg}{\textbf{DPL~(Ours)}}  & \cellcolor{mygray-bg}{} & \cellcolor{mygray-bg}{\textcolor{blue}{25.42}} & \cellcolor{mygray-bg}{\textcolor{blue}{29.99}} & \cellcolor{mygray-bg}{\textcolor{blue}{31.85}} & \cellcolor{mygray-bg}{\textcolor{blue}{26.15}} & \cellcolor{mygray-bg}{\textcolor{blue}{30.37}} &\cellcolor{mygray-bg}{\textcolor{blue}{32.13}}  \\
   &  & \cellcolor{mygray-bg}{\textbf{DPL\textsuperscript{$\dagger$}~(Ours)}}  & \cellcolor{mygray-bg}{} & \cellcolor{mygray-bg}{\textbf{\textcolor{red}{26.02}}} & \cellcolor{mygray-bg}{\textbf{\textcolor{red}{30.50}}} & \cellcolor{mygray-bg}{\textcolor{red}{\textbf{32.33}}} & \cellcolor{mygray-bg}{\textcolor{red}{\textbf{26.44}}} & \cellcolor{mygray-bg}{\textcolor{red}{\textbf{30.81}}} & \cellcolor{mygray-bg}{\textcolor{red}{\textbf{32.67}}} \\
   \cline{2-10}
   & \multirow{11}{*}{5-shot} & RelNet~\cite{sung2018learning} & $_{\textit{CVPR'18}}$  &  20.29 & 26.56 & 29.70 & 15.72 & 20.66 & 23.33  \\
   & & ProtoNet~\cite{snell2017prototypical} & $_{\textit{NIPS'17}}$  &  25.00 & 30.18 & 32.77 & 20.28 & 24.65 & 27.00\\
    & & LPaF~\cite{dornadula2019visual} & $_{\textit{CVPR'19}}$  &  -- & 20.90 & -- & -- & -- & -- \\
    &  & LKMN~\cite{li2022zero} & $_{\textit{TMM'22}}$  &  -- & 22.00 & -- & -- & -- & -- \\
    & & VinVL~\cite{zhang2021vinvl} & $_{\textit{CVPR'21}}$ & 24.80 &  30.20 &  33.10  &  26.40  &  31.40  &  34.10  \\
   &  & CoOp~\cite{zhou2022learning} & $_{\textit{IJCV'22}}$  &  29.03 & 34.60 & 37.20 & 29.12 & 33.81 & 36.40 \\
   & &  PENET~\cite{zheng2023prototype}  &  $_{\textit{CVPR'23}}$   &  31.89   &  37.64   &   40.10 &   30.52  &  35.69   &  37.95  \\
   &  & KFV~\cite{yu2023knowledge} & $_{\textit{Arxiv'23}}$ & 33.20 &  40.30 &  43.60 &  33.10 &  40.20  &  43.70 \\
     & & TCP~\cite{yao2024tcp} & $_{\textit{CVPR'24}}$ & 29.97 &  37.18 & 39.97 & 27.63  &  34.05  & 36.52  \\
     
   &  & \cellcolor{mygray-bg}{\textbf{DPL~(Ours)}}  &  \cellcolor{mygray-bg}{} & \cellcolor{mygray-bg}{\textcolor{blue}{37.07}} & \cellcolor{mygray-bg}{\textcolor{blue}{43.30}} & \cellcolor{mygray-bg}{\textcolor{blue}{46.08}} &  \cellcolor{mygray-bg}{\textcolor{blue}{36.24}} & \cellcolor{mygray-bg}{\textcolor{blue}{42.07}} & \cellcolor{mygray-bg}{\textcolor{blue}{44.67}}  \\
   &  & \cellcolor{mygray-bg}{\textbf{DPL\textsuperscript{$\dagger$}~(Ours)}}   &  \cellcolor{mygray-bg}{} & \cellcolor{mygray-bg}{\textbf{\textcolor{red}{38.46}}} & \cellcolor{mygray-bg}{\textbf{\textcolor{red}{44.89}}} & \cellcolor{mygray-bg}{\textbf{\textcolor{red}{47.73}}} & \cellcolor{mygray-bg}{\textcolor{red}{\textbf{37.07}}} & \cellcolor{mygray-bg}{\textcolor{red}{\textbf{43.28}}} & \cellcolor{mygray-bg}{\textcolor{red}{\textbf{45.87}}} \\
   \cline{2-10}
   &  \multirow{9}{*}{10-shot}& RelNet~\cite{sung2018learning} & $_{\textit{CVPR'18}}$  &  20.08 & 27.26 & 31.22 & 15.41 & 21.97 & 25.41 \\
    & & ProtoNet~\cite{snell2017prototypical} & $_{\textit{NIPS'17}}$  &  25.15 & 32.72 & 36.30 & 19.20 & 24.52 & 27.76 \\
    & &  VinVL~\cite{zhang2021vinvl}  &  $_{\textit{CVPR'21}}$  &  29.20  &  35.10  &  38.40  &  31.00 & 36.90 &  40.60  \\
   &  & CoOp~\cite{zhou2022learning} & $_{\textit{IJCV'22}}$  &  30.42 & 36.83 & 39.95 & 33.07 & 39.62 & 42.36 \\
   & &  PENET~\cite{zheng2023prototype}  &  $_{\textit{CVPR'23}}$   &  38.20    &  44.28   &  46.54   &   36.44  &  41.97   & 44.16 \\
   &  & KFV~\cite{yu2023knowledge} & $_{\textit{Arxiv'23}}$ & 37.80 &  44.30 &  48.30 &  38.50 & 45.50 &  \textcolor{blue}{49.50} \\
    & & TCP~\cite{yao2024tcp} & $_{\textit{CVPR'24}}$ &  33.81 & 41.48  & 44.87 & 33.95  &  40.86  &  44.06 \\ 
    
   &  & \cellcolor{mygray-bg}{\textbf{DPL~(Ours)}}  &  \cellcolor{mygray-bg}{} & \cellcolor{mygray-bg}{\textcolor{blue}{40.31}} & \cellcolor{mygray-bg}{\textcolor{blue}{47.02}} & \cellcolor{mygray-bg}{\textcolor{blue}{49.53}} & \cellcolor{mygray-bg}{\textcolor{blue}{39.35}} & \cellcolor{mygray-bg}{\textcolor{blue}{45.66}} & \cellcolor{mygray-bg}{48.15} \\
   &  & \cellcolor{mygray-bg}{\textbf{DPL\textsuperscript{$\dagger$}~(Ours)}}   & \cellcolor{mygray-bg}{} & \cellcolor{mygray-bg}{\textbf{\textcolor{red}{41.05}}} & \cellcolor{mygray-bg}{\textbf{\textcolor{red}{47.87}}} & \cellcolor{mygray-bg}{\textbf{\textcolor{red}{50.64}}} & \cellcolor{mygray-bg}{\textbf{\textcolor{red}{41.03}}} & \cellcolor{mygray-bg}{\textbf{\textcolor{red}{48.00}}} & \cellcolor{mygray-bg}{\textbf{\textcolor{red}{50.60}}}  \\
\hline\thickhline
    \end{tabular}
    }
    \label{tab:VG-25-predcls-unseen}
\end{table}

\begin{table}[t]
    \centering
    \caption{Unseen predicate performance (\%) comparison of different methods on GQA-50 dataset \wrt PredCls with K-shot. The \textcolor{red}{\textbf{best}} and \textcolor{blue}{second best} results are marked according to formats.}
    \vspace{-1em}
    \scalebox{0.9}{
    \begin{tabular}{ c |c |l l | c c c | c c c}
    \hline\thickhline
    \multirow{2}{*}{D} & \multirow{2}{*}{K-shot} & \multicolumn{2}{c|}{\multirow{2}{*}{Method}} & \multicolumn{3}{c|}{Recall} &  \multicolumn{3}{c}{meanRecall} \\
     \cline{5-10}
      & & & & @20 & @50 & @100  & @20 & @50 & @100  \\
  \hline\hline
\multirow{18}{*}{\begin{sideways}GQA-50\end{sideways}}& \multirow{7}{*}{1-Shot} 
& RelNet~\cite{sung2018learning} & $_{\textit{CVPR'18}}$   & 11.36  &  16.14 & 17.97  & 9.73 &  14.03 & 16.06 \\
 &  & ProtoNet~\cite{snell2017prototypical} & $_{\textit{NIPS'17}}$& 21.37  & 25.37  & 27.42  &20.36 & 24.70  & 26.86   \\
& & CoOp~\cite{zhou2022learning} & $_{\textit{IJCV'22}}$ & 22.10 &  24.00  & 24.74  & 21.16  & 23.25 &  23.97 \\
& & PENET~\cite{zheng2023prototype} & $_{\textit{CVPR'23}}$ & 26.49  & 29.06  & 30.04  & 25.24 & 28.25  & 29.41 \\
& &  TCP~\cite{yao2024tcp}  &  $_{\textit{CVPR'24}}$  &  24.23   &  26.88  &  27.96  &  23.95   & 26.91    &  28.05   \\
 &  & \cellcolor{mygray-bg}{\textbf{DPL~(Ours)}}   & \cellcolor{mygray-bg}{}& \cellcolor{mygray-bg}{\textcolor{blue}{28.23}}  & \cellcolor{mygray-bg}{\textcolor{blue}{30.99}}  & \cellcolor{mygray-bg}{\textcolor{blue}{31.96}} & \cellcolor{mygray-bg}{\textcolor{blue}{27.13}}  & \cellcolor{mygray-bg}{\textcolor{blue}{30.08}} & \cellcolor{mygray-bg}{\textcolor{blue}{31.27}}  \\
&  & \cellcolor{mygray-bg}{\textbf{DPL\textsuperscript{$\dagger$}~(Ours)}}   & \cellcolor{mygray-bg}{} & \cellcolor{mygray-bg}{\textcolor{red}{\textbf{30.96}}}  & \cellcolor{mygray-bg}{\textcolor{red}{\textbf{34.04}}}  & \cellcolor{mygray-bg}{\textcolor{red}{\textbf{35.15}}} &  \cellcolor{mygray-bg}{\textcolor{red}{\textbf{28.45}}} & \cellcolor{mygray-bg}{\textcolor{red}{\textbf{31.98}}} &  \cellcolor{mygray-bg}{\textcolor{red}{\textbf{33.23}}} \\
\cline{2-10}
\cline{2-8}
 &  \multirow{7}{*}{5-Shot} 
& RelNet~\cite{sung2018learning} & $_{\textit{CVPR'18}}$  & 18.21  & 21.65  & 23.17 & 17.43   & 21.18 & 22.85  \\
 &  & ProtoNet~\cite{snell2017prototypical} & $_{\textit{NIPS'17}}$  & 27.36  & 31.23  & 33.01 & 27.88  & 32.10 & 34.05  \\
&  & CoOp~\cite{zhou2022learning} & $_{\textit{IJCV'22}}$ & 34.96 & 37.61  & 38.90 & 35.40 & 38.29 & 39.66  \\
& & PENET~\cite{zheng2023prototype} & $_{\textit{CVPR'23}}$  & 36.73  & 39.59  & 40.60 &  \textcolor{blue}{37.58} & \textcolor{blue}{40.87} & \textcolor{blue}{42.14}  \\
& & TCP~\cite{yao2024tcp} & $_{\textit{CVPR'24}}$ & \textcolor{blue}{36.84}  & \textcolor{blue}{40.11}  & 41.44 & 36.61  & 40.43   & 41.81   \\

&  & \cellcolor{mygray-bg}{\textbf{DPL~(Ours)}}   & \cellcolor{mygray-bg}{} & \cellcolor{mygray-bg}{35.61}  & \cellcolor{mygray-bg}{39.91}  & \cellcolor{mygray-bg}{\textcolor{blue}{41.77}} & \cellcolor{mygray-bg}{35.39}  & \cellcolor{mygray-bg}{40.12} & \cellcolor{mygray-bg}{42.09}  \\
&  & \cellcolor{mygray-bg}{\textbf{DPL\textsuperscript{$\dagger$}~(Ours)}}  & \cellcolor{mygray-bg}{} &  \cellcolor{mygray-bg}{\textcolor{red}{\textbf{39.74}}} & \cellcolor{mygray-bg}{\textcolor{red}{\textbf{44.45}}}  &\cellcolor{mygray-bg}{\textcolor{red}{\textbf{46.58}}}  & \cellcolor{mygray-bg}{\textcolor{red}{\textbf{39.63}}}  & \cellcolor{mygray-bg}{\textcolor{red}{\textbf{45.18}}} & \cellcolor{mygray-bg}{\textcolor{red}{\textbf{47.63}}}  \\
\cline{2-10}
& \multirow{7}{*}{10-Shot} 
& RelNet~\cite{sung2018learning} & $_{\textit{CVPR'18}}$  & 19.18  & 22.61  & 23.91 & 19.61  & 23.60 & 25.04  \\
& & ProtoNet~\cite{snell2017prototypical} & $_{\textit{NIPS'17}}$    & 28.71   & 32.50  & 34.16 & 29.97  & 34.30  & 36.04  \\
&  & CoOp~\cite{zhou2022learning} & $_{\textit{IJCV'22}}$  & 39.09 & 42.47  & 43.86 & 38.61  &42.04 & 43.62  \\
& & PENET~\cite{zheng2023prototype} & $_{\textit{CVPR'23}}$  &  \textcolor{blue}{40.36} & \textcolor{blue}{43.50} & \textcolor{blue}{44.62} & \textcolor{blue}{40.35}  & 43.82 & 45.03 \\
& &  TCP~\cite{yao2024tcp}  & $_{\textit{CVPR'24}}$  &  37.92  &  41.61  &  43.03 &  39.71  & \textcolor{blue}{44.00} & \textcolor{blue}{45.68} \\
&  & \cellcolor{mygray-bg}{\textbf{DPL~(Ours)}}   & \cellcolor{mygray-bg}{} & \cellcolor{mygray-bg}{38.13}  & \cellcolor{mygray-bg}{41.94}  & \cellcolor{mygray-bg}{43.74} & \cellcolor{mygray-bg}{37.16}  & \cellcolor{mygray-bg}{41.24} & \cellcolor{mygray-bg}{43.33}  \\
& & \cellcolor{mygray-bg}{\textbf{DPL\textsuperscript{$\dagger$}~(Ours)}}   & \cellcolor{mygray-bg}{} &  \cellcolor{mygray-bg}{\textcolor{red}{\textbf{41.55}}} & \cellcolor{mygray-bg}{\textcolor{red}{\textbf{46.93}}}  & \cellcolor{mygray-bg}{\textcolor{red}{\textbf{49.14}}} & \cellcolor{mygray-bg}{\textcolor{red}{\textbf{41.70}}}  & \cellcolor{mygray-bg}{\textcolor{red}{\textbf{47.85}}} & \cellcolor{mygray-bg}{\textcolor{red}{\textbf{50.18}}}  \\

   \hline\thickhline
    \end{tabular}
    }
    \label{tab:GQA50_predcls}
\end{table}

\subsection{Performance Comparison with Baselines}
Following the previous work~\cite{yu2023knowledge,li2022zero}, we mainly compare the performance of unseen predicates for different models on the PredCls setting. We also report model performance on the SGCls setting and performance on seen predicates for a complete comparison.
Our experiments and discussion on different models is organized as follows:

To compare model performance on seen predicates, we conduct experiments on the PredCls setting based with VG-25, VG-60 and GQA-50 datasets respectively and report their performance on both Recall and meanRecall as shown in Table~\ref{tab:VG-25-predcls-unseen},~\ref{tab:GQA50_predcls} and ~\ref{tab:VG60_predcls}. 
For complete comparison, we also conduct experiments on the SGCls setting and report unseen predicate performance on VG-25 and VG-60 as shown in Table~\ref{tab:VG-25-unseen-sgcls} and~\ref{tab:VG-60-unseen-sgcls}.

In addition, to observe model performance on seen predicates, we report seen predicate performance on both PredCls and SGCls based on VG-25 dataset as shown in Table~\ref{tab:VG25-seen-predcls} and~\ref{tab:VG25-seen-sgcls}.

\subsubsection{\textbf{Unseen Predicate Performance}}
We report unseen predicate category performance of the model \wrt PredCls on three different datasets in Table~\ref{tab:VG-25-predcls-unseen}, ~\ref{tab:GQA50_predcls} and ~\ref{tab:VG60_predcls} respectively.

\begin{table}[t]
    \centering
    \caption{Unseen predicate performance (\%) comparison of different methods on VG-60 dataset\wrt PredCls with K-shot. The \textcolor{red}{\textbf{best}} and \textcolor{blue}{second best} results are marked according to formats.}
    \vspace{-1em}
    \scalebox{0.90}{
    \begin{tabular}{ c |c |l l | c c c | c c c}
    \hline\thickhline
    \multirow{2}{*}{D} & \multirow{2}{*}{K-shot} & \multicolumn{2}{c|}{\multirow{2}{*}{Method}} & \multicolumn{3}{c|}{Recall} &  \multicolumn{3}{c}{meanRecall} \\
     \cline{5-10}
      & & & & @20 & @50 & @100  & @20 & @50 & @100  \\
  \hline\hline
\multirow{15}{*}{\begin{sideways}VG-60\end{sideways}} & \multirow{5}{*}{1-Shot} 
& RelNet~\cite{sung2018learning} & $_{\textit{CVPR'18}}$  & 7.34 & 10.22 & 11.92 & 6.68 & 9.56 & 11.40 \\
 & & ProtoNet~\cite{snell2017prototypical} & $_{\textit{NIPS'17}}$ & 10.44 & 14.12 & 16.49 & 10.00 & 13.83 & 16.18 \\
& & CoOp~\cite{zhou2022learning} & $_{\textit{IJCV'22}}$ & \textcolor{blue}{17.48}  & \textcolor{blue}{21.09}  & 22.92  & \textcolor{blue}{17.02}  & \textcolor{blue}{20.70} & \textcolor{blue}{22.52}  \\
 &  & \cellcolor{mygray-bg}{\textbf{DPL~(Ours)}} &\cellcolor{mygray-bg}{} & \cellcolor{mygray-bg}{16.81} & \cellcolor{mygray-bg}{21.06} & \cellcolor{mygray-bg}{\textcolor{blue}{23.10}} & \cellcolor{mygray-bg}{16.27} & \cellcolor{mygray-bg}{20.26} & \cellcolor{mygray-bg}{22.22} \\
&  & \cellcolor{mygray-bg}{\textbf{DPL\textsuperscript{$\dagger$}~(Ours)}}  & \cellcolor{mygray-bg}{} & \cellcolor{mygray-bg}{\textbf{\textcolor{red}{18.24}}} & \cellcolor{mygray-bg}{\textbf{\textcolor{red}{22.58}}} & \cellcolor{mygray-bg}{\textbf{\textcolor{red}{24.67}}} & \cellcolor{mygray-bg}{\textbf{\textcolor{red}{17.80}}} & \cellcolor{mygray-bg}{\textbf{\textcolor{red}{21.86}}} & \cellcolor{mygray-bg}{\textbf{\textcolor{red}{23.87}}}  \\
\cline{2-10}
 &  \multirow{5}{*}{5-Shot} 
& RelNet~\cite{sung2018learning} & $_{\textit{CVPR'18}}$ & 12.76 & 17.06 & 19.42 & 12.16 & 16.88	 & 19.48 \\
 &  & ProtoNet~\cite{snell2017prototypical} & $_{\textit{NIPS'17}}$ & 17.43 & 22.51	 & 24.94 & 16.60 & 21.97 & 24.51 \\
&  & CoOp~\cite{zhou2022learning} & $_{\textit{IJCV'22}}$ & 25.79 & 30.97   & 33.38 & 25.48 &  30.70 & 33.19 \\
&  & \cellcolor{mygray-bg}{\textbf{DPL~(Ours)}}  & \cellcolor{mygray-bg}{} & \cellcolor{mygray-bg}{\textcolor{blue}{26.40}} & \cellcolor{mygray-bg}{\textcolor{blue}{33.10}} & \cellcolor{mygray-bg}{\textcolor{blue}{35.80}} & \cellcolor{mygray-bg}{\textcolor{blue}{25.87}} & \cellcolor{mygray-bg}{\textcolor{blue}{32.39}} & \cellcolor{mygray-bg}{\textcolor{blue}{35.14}} \\
&  & \cellcolor{mygray-bg}{\textbf{DPL\textsuperscript{$\dagger$}~(Ours)}}  & \cellcolor{mygray-bg}{} & \cellcolor{mygray-bg}{\textcolor{red}{\textbf{29.32}}} & \cellcolor{mygray-bg}{\textbf{\textcolor{red}{36.27}}} & \cellcolor{mygray-bg}{\textbf{\textcolor{red}{39.19}}} & \cellcolor{mygray-bg}{\textbf{\textcolor{red}{28.25}}} & \cellcolor{mygray-bg}{\textbf{\textcolor{red}{35.30}}} & \cellcolor{mygray-bg}{\textbf{\textcolor{red}{38.30}}}  \\
\cline{2-10}
& \multirow{5}{*}{10-Shot} 
& RelNet~\cite{sung2018learning} & $_{\textit{CVPR'18}}$ & 14.47 & 20.27 & 23.18 & 13.57 & 19.64 & 22.66 \\
& & ProtoNet~\cite{snell2017prototypical} & $_{\textit{NIPS'17}}$   & 19.58 & 25.00 & 27.95 & 18.44 & 24.20 & 27.31  \\
&  & CoOp~\cite{zhou2022learning} & $_{\textit{IJCV'22}}$ & \textcolor{blue}{28.35}  &  34.59 & 37.13  & \textcolor{blue}{28.23}  & 34.35 & 36.76 \\
&  & \cellcolor{mygray-bg}{\textbf{DPL~(Ours)}}  & \cellcolor{mygray-bg}{} & \cellcolor{mygray-bg}{28.14}  &  \cellcolor{mygray-bg}{\textcolor{blue}{34.75}}  &  \cellcolor{mygray-bg}{\textcolor{blue}{38.08}}    &  \cellcolor{mygray-bg}{27.56}  &  \cellcolor{mygray-bg}{\textcolor{blue}{34.36}}  &  \cellcolor{mygray-bg}{\textcolor{blue}{37.78}} \\
& & \cellcolor{mygray-bg}{\textbf{DPL\textsuperscript{$\dagger$}~(Ours)}}  &\cellcolor{mygray-bg}{} & \cellcolor{mygray-bg}{\textbf{\textcolor{red}{30.57}}} &  \cellcolor{mygray-bg}{\textbf{\textcolor{red}{38.36}}} &  \cellcolor{mygray-bg}{\textbf{\textcolor{red}{41.86}}} &  \cellcolor{mygray-bg}{\textbf{\textcolor{red}{30.17}}} &  \cellcolor{mygray-bg}{\textbf{\textcolor{red}{37.96}}} &  \cellcolor{mygray-bg}{\textbf{\textcolor{red}{41.54}}}  \\

   \hline\thickhline
    \end{tabular}
    }
    \label{tab:VG60_predcls}
\end{table}

\paragraph{\textbf{Performance Comparison on VG-25}} We report unseen predicate performance on Table~\ref{tab:VG-25-predcls-unseen}. From the table, we can see that our method has achieved state-of-the-art performance on both Recall@N and mean Recall@N metrics in all K-shot settings for unseen predicate categories. And due to the fine-tuning process on the support samples of unseen categories, the performance of \emph{DPL\textsuperscript{$\dagger$}} has been significantly improved compared to the performance of \emph{DPL}. In addition, we can see that although our \emph{DPL} has not fine-tuned on the unseen categories, its performance is still ahead of other baseline models. Especially in the most challenging $1$-shot setting, both our \emph{DPL} and \emph{DPL\textsuperscript{$\dagger$}} have shown great advantages compared to baseline methods.\\

\begin{table}[t]
    \centering
    \caption{Unseen predicate performance (\%) comparison of different methods on VG-25 dataset \wrt SGCls with K-shot. The \textcolor{red}{\textbf{best}} and \textcolor{blue}{second best} results are marked according to formats.}
    \vspace{-1em}
    \scalebox{0.9}{
    \begin{tabular}{ c |c | l l | c c c | c c c}
    \hline\thickhline
    \multirow{2}{*}{D} & \multirow{2}{*}{K-shot} & \multicolumn{2}{c|}{\multirow{2}{*}{Method}} & \multicolumn{3}{c|}{Recall} &  \multicolumn{3}{c}{meanRecall} \\
     \cline{5-10}
      & & & & @20 & @50 & @100  & @20 & @50 & @100  \\
      \hline
      \hline
     \multirow{15}{*}{\begin{sideways}VG-25\end{sideways}}  &  \multirow{5}{*}{1-Shot} &
RelNet~\cite{sung2018learning} & $_{\textit{CVPR'18}}$  &  7.16 & 9.56 & 10.40	& 5.92 & 7.64 & 8.40\\
   & & ProtoNet~\cite{snell2017prototypical} & $_{\textit{NIPS'17}}$   & 5.41 & 7.50 & 8.70 & 5.01 & 6.46 & 7.34\\
   & & CoOp~\cite{zhou2022learning} & $_{\textit{IJCV'22}}$  &  8.23 & 9.49 & 10.06  & 7.40 & 8.45 & 9.08\\
   & & \cellcolor{mygray-bg}{\textbf{DPL~(Ours)}}   &\cellcolor{mygray-bg}{} & \cellcolor{mygray-bg}{\textcolor{blue}{8.69}} & \cellcolor{mygray-bg}{\textcolor{blue}{10.24}} & \cellcolor{mygray-bg}{\textcolor{blue}{10.95}}  & \cellcolor{mygray-bg}{\textcolor{blue}{10.41}} & \cellcolor{mygray-bg}{\textcolor{blue}{11.95}} & \cellcolor{mygray-bg}{\textcolor{blue}{12.58}} \\
   & & \cellcolor{mygray-bg}{\textbf{DPL\textsuperscript{$\dagger$}~(Ours)}}   & \cellcolor{mygray-bg}{}  & \cellcolor{mygray-bg}{\textcolor{red}{\textbf{11.66}}} & \cellcolor{mygray-bg}{\textcolor{red}{\textbf{13.64}}} & \cellcolor{mygray-bg}{\textcolor{red}{\textbf{14.34}}} & \cellcolor{mygray-bg}{\textcolor{red}{\textbf{13.95}}} & \cellcolor{mygray-bg}{\textcolor{red}{\textbf{15.66}}} & \cellcolor{mygray-bg}{\textcolor{red}{\textbf{16.27}}}\\
   \cline{2-10}
   & \multirow{5}{*}{5-shot}& RelNet~\cite{sung2018learning} & $_{\textit{CVPR'18}}$   &  7.99 & 10.23 & 11.04  & 6.95 & 8.80 & 9.61 \\
   & & ProtoNet~\cite{snell2017prototypical} & $_{\textit{NIPS'17}}$   &  10.51 & 12.44 & 13.16 & 9.02 & 10.82 & 11.51\\
   & & CoOp~\cite{zhou2022learning} & $_{\textit{IJCV'22}}$  & 11.80 & 14.07 & 14.98 &  10.57 & 12.47 & 13.25\\
   & & \cellcolor{mygray-bg}{\textbf{DPL~(Ours)}}   &\cellcolor{mygray-bg}{} &  \cellcolor{mygray-bg}{\textcolor{blue}{13.21}} & \cellcolor{mygray-bg}{\textcolor{blue}{15.62}} & \cellcolor{mygray-bg}{\textcolor{blue}{16.66}} & \cellcolor{mygray-bg}{\textcolor{blue}{16.54}} & \cellcolor{mygray-bg}{\textcolor{blue}{18.70}} & \cellcolor{mygray-bg}{\textcolor{blue}{19.80}} \\
   & & \cellcolor{mygray-bg}{\textbf{DPL\textsuperscript{$\dagger$}~(Ours)}}  & \cellcolor{mygray-bg}{}  &  \cellcolor{mygray-bg}{\textcolor{red}{\textbf{17.38}}} & \cellcolor{mygray-bg}{\textcolor{red}{\textbf{19.80}}} & \cellcolor{mygray-bg}{\textcolor{red}{\textbf{20.69}}}  & \cellcolor{mygray-bg}{\textcolor{red}{\textbf{18.93}}} & \cellcolor{mygray-bg}{\textcolor{red}{\textbf{21.55}}} & \cellcolor{mygray-bg}{\textcolor{red}{\textbf{22.43}}}\\
   \cline{2-10}
   & \multirow{5}{*}{10-shot}& RelNet~\cite{sung2018learning} & $_{\textit{CVPR'18}}$  & 10.37 & 12.31 & 12.92 & 	8.23 & 10.05 & 10.78\\
   & & ProtoNet~\cite{snell2017prototypical} & $_{\textit{NIPS'17}}$   & 11.68 & 13.77 & 14.56 & 9.60 & 11.77 & 12.50\\
   & & CoOp~\cite{zhou2022learning} & $_{\textit{IJCV'22}}$  &  13.43 & 16.16 & 17.31 & 14.16 & 16.28 & 17.12\\
   & & \cellcolor{mygray-bg}{\textbf{DPL~(Ours)}}  & \cellcolor{mygray-bg}{} &  \cellcolor{mygray-bg}{\textcolor{blue}{14.30}} & \cellcolor{mygray-bg}{\textcolor{blue}{16.81}} & \cellcolor{mygray-bg}{\textcolor{blue}{17.69}} & \cellcolor{mygray-bg}{\textcolor{blue}{17.04}} & \cellcolor{mygray-bg}{\textcolor{blue}{19.23}} & \cellcolor{mygray-bg}{\textcolor{blue}{20.19}} \\
   & & \cellcolor{mygray-bg}{\textbf{DPL\textsuperscript{$\dagger$}~(Ours)}}  & \cellcolor{mygray-bg}{}  & \cellcolor{mygray-bg}{\textcolor{red}{\textbf{18.18}}} & \cellcolor{mygray-bg}{\textcolor{red}{\textbf{20.87}}} & \cellcolor{mygray-bg}{\textcolor{red}{\textbf{21.88}}} & \cellcolor{mygray-bg}{\textcolor{red}{\textbf{19.84}}} & \cellcolor{mygray-bg}{\textcolor{red}{\textbf{22.49}}} & \cellcolor{mygray-bg}{\textcolor{red}{\textbf{23.50}}} \\
     \hline\thickhline
  \end{tabular}}
  \label{tab:VG-25-unseen-sgcls}
\end{table}

\begin{table}[]
    \centering
    \caption{Unseen predicate performance (\%) comparison of different methods on VG-60 dataset \wrt SGCls with K-shot. The \textcolor{red}{\textbf{best}} and \textcolor{blue}{second best} results are marked according to formats.}
    \vspace{-1em}
    \scalebox{0.9}{
   \begin{tabular}{ c |c | l l | c c c | c c c}
    \hline\thickhline
    \multirow{2}{*}{D} & \multirow{2}{*}{K-shot} & \multicolumn{2}{c|}{\multirow{2}{*}{Method}} & \multicolumn{3}{c|}{Recall} &  \multicolumn{3}{c}{meanRecall} \\
     \cline{5-10}
      & & & & @20 & @50 & @100  & @20 & @50 & @100  \\
      \hline
      \hline
\multirow{15}{*}{\begin{sideways}VG-60\end{sideways}} & \multirow{5}{*}{1-Shot} 
& RelNet~\cite{sung2018learning} & $_{\textit{CVPR'18}}$  & 3.44 & 4.30 & 4.68 & 3.34 & 4.38 & 4.79 \\
 & & ProtoNet~\cite{snell2017prototypical} & $_{\textit{NIPS'17}}$ & 3.98 & 5.18 & 5.84 & 3.76 & 4.87 & 5.50\\
& & CoOp~\cite{zhou2022learning} & $_{\textit{IJCV'22}}$ & 5.93 & 6.81 & 7.29 & 5.80 & 6.66 & 7.16\\
 & & \cellcolor{mygray-bg}{\textbf{DPL~(Ours)}} & \cellcolor{mygray-bg}{} &  \cellcolor{mygray-bg}{\textcolor{blue}{7.33}} &  \cellcolor{mygray-bg}{\textcolor{blue}{8.70}} &  \cellcolor{mygray-bg}{\textcolor{blue}{9.18}} &  \cellcolor{mygray-bg}{\textcolor{blue}{7.34}} &  \cellcolor{mygray-bg}{\textcolor{red}{\textbf{8.60}}} &  \cellcolor{mygray-bg}{\textcolor{blue}{9.06}} \\
& &  \cellcolor{mygray-bg}{\textbf{DPL\textsuperscript{$\dagger$}~(Ours)}} & \cellcolor{mygray-bg}{} &  \cellcolor{mygray-bg}{\textcolor{red}{\textbf{7.51}}} &  \cellcolor{mygray-bg}{\textcolor{red}{\textbf{8.78}}} &  \cellcolor{mygray-bg}{\textcolor{red}{\textbf{9.33}}} &  \cellcolor{mygray-bg}{\textcolor{red}{\textbf{7.40}}} &  \cellcolor{mygray-bg}{\textcolor{blue}{8.58}} &  \cellcolor{mygray-bg}{\textcolor{red}{\textbf{9.10}}}  \\
\cline{2-10}
 & \multirow{5}{*}{5-Shot} 
& RelNet~\cite{sung2018learning} & $_{\textit{CVPR'18}}$ & 5.27 & 6.49 & 6.89 & 5.08 & 6.57 & 7.03 \\
 & & ProtoNet~\cite{snell2017prototypical} & $_{\textit{NIPS'17}}$ & 6.73 & 8.30 & 9.03 & 6.31 & 7.88 & 8.56 \\
& & CoOp~\cite{zhou2022learning} & $_{\textit{IJCV'22}}$ & 9.39 & 10.94 & 11.68 & 9.29 & 10.84 & 11.56\\
& &  \cellcolor{mygray-bg}{\textbf{DPL~(Ours)}} &  \cellcolor{mygray-bg}{} &   \cellcolor{mygray-bg}{\textcolor{blue}{10.76}} &  \cellcolor{mygray-bg}{\textcolor{blue}{12.61}} &  \cellcolor{mygray-bg}{\textcolor{blue}{13.34}} &    \cellcolor{mygray-bg}{\textcolor{blue}{10.53}} &  \cellcolor{mygray-bg}{\textcolor{blue}{12.31}} &  \cellcolor{mygray-bg}{\textcolor{blue}{13.06}} \\
& &  \cellcolor{mygray-bg}{\textbf{DPL\textsuperscript{$\dagger$}~(Ours)}} &  \cellcolor{mygray-bg}{}  &   \cellcolor{mygray-bg}{\textcolor{red}{\textbf{11.83}}} &  \cellcolor{mygray-bg}{\textcolor{red}{\textbf{13.98}}} &  \cellcolor{mygray-bg}{\textcolor{red}{\textbf{15.05}}} &  \cellcolor{mygray-bg}{\textcolor{red}{\textbf{11.27}}} &  \cellcolor{mygray-bg}{\textcolor{red}{\textbf{13.41}}} &  \cellcolor{mygray-bg}{\textcolor{red}{\textbf{14.45}}}  \\
\cline{2-10}
& \multirow{5}{*}{10-Shot} 
& RelNet~\cite{sung2018learning} & $_{\textit{CVPR'18}}$ & 5.76 & 6.86 & 7.34 & 5.68 & 6.83  & 7.33 \\
& & ProtoNet~\cite{snell2017prototypical} & $_{\textit{NIPS'17}}$   & 7.55 & 8.91 & 9.50 & 7.10 & 8.40 & 8.98 \\
& & CoOp~\cite{zhou2022learning} & $_{\textit{IJCV'22}}$ & \textcolor{blue}{11.41} & \textcolor{blue}{13.04} & 13.77 & \textcolor{blue}{11.06} & 12.70 & 13.37 \\
&  &  \cellcolor{mygray-bg}{\textbf{DPL~(Ours)}} & \cellcolor{mygray-bg}{} &   \cellcolor{mygray-bg}{10.98} &  \cellcolor{mygray-bg}{12.92} &  \cellcolor{mygray-bg}{\textcolor{blue}{13.80}} &  \cellcolor{mygray-bg}{10.89} &  \cellcolor{mygray-bg}{\textcolor{blue}{12.86}} &  \cellcolor{mygray-bg}{\textcolor{blue}{13.75}} \\
& &  \cellcolor{mygray-bg}{\textbf{DPL\textsuperscript{$\dagger$}~(Ours)}} & \cellcolor{mygray-bg}{} &   \cellcolor{mygray-bg}{\textcolor{red}{\textbf{11.98}}} &   \cellcolor{mygray-bg}{\textcolor{red}{\textbf{14.52}}} &   \cellcolor{mygray-bg}{\textcolor{red}{\textbf{15.53}}} &   \cellcolor{mygray-bg}{\textcolor{red}{\textbf{11.60}}} &   \cellcolor{mygray-bg}{\textcolor{red}{\textbf{14.06}}} &   \cellcolor{mygray-bg}{\textcolor{red}{\textbf{15.06}}}  \\
  \hline\thickhline
  \end{tabular}}
  \label{tab:VG-60-unseen-sgcls}
\end{table}

The performance of \emph{ProtoNet} and \emph{RelNet} has been greatly improved compared to the models \emph{LPaF} and \emph{LKMN}. This is mainly due to the excellent representation ability of the metric learning method in few-shot settings. However, compared with our \emph{DPL} model, the performance of \emph{ProtoNet} and \emph{RelNet} is slightly inferior. This is mainly because these methods model the prototype representation of each category only based on knowledge in the support samples without exploiting the additional knowledge.
Meanwhile, we can observe that the performance of the Vision-Language based methods, \emph{CoOp}, \emph{TCP} and \emph{VinVL}, is better than \emph{ProtoNet} and \emph{RelNet}. This is because they introduce the knowledge of powerful VL models and exploit a prompt learning approach to quickly transfer the knowledge of the VL model. 
However, these methods perform worse than \emph{KFV} and \emph{PENET}, which are tailored to visual relationship triplet recognition by exploiting the compositional knowledge of subject and object pairs.
Nevertheless, our methods still have achieved the most excellent performance compared to all baselines.
This is because our method represents different predicate categories with multiple prototypes based on subject and objects, which effectively extracts the semantic knowledge of relationship triplets hidden in the VL model, thereby improving the knowledge transferring ability on the unseen predicate categories.

\begin{table}[t]
    \centering
    \caption{Seen predicate performance (\%) comparison of different methods on VG-25 dataset \wrt PredCls with K-shot. The \textcolor{red}{\textbf{best}} and \textcolor{blue}{second best} results are marked according to formats.}
    \vspace{-1em}
    \scalebox{0.9}{
    \begin{tabular}{ c |c | l l | c c c | c c c}
    \hline\thickhline
    \multirow{2}{*}{D} & \multirow{2}{*}{K-shot} & \multicolumn{2}{c|}{\multirow{2}{*}{Method}} & \multicolumn{3}{c|}{Recall} &  \multicolumn{3}{c}{meanRecall} \\
     \cline{5-10}
      & & & & @20 & @50 & @100  & @20 & @50 & @100  \\
      \hline
      \hline
   \multirow{15}{*}{\begin{sideways}VG-25\end{sideways}} & \multirow{5}{*}{1-shot}& RelNet~\cite{sung2018learning} & $_{\textit{CVPR'18}}$  & 21.56 & 30.15 & 34.37 & 15.53 & 21.81 & 25.12 \\
    & & ProtoNet~\cite{snell2017prototypical} &  $_{\textit{NIPS'17}}$ & 15.21 & 21.98 & 25.33 & 13.11 & 18.54 & 21.62  \\
    & & CoOp~\cite{zhou2022learning}  & $_{\textit{IJCV'22}}$  & 29.27 & 37.32 & 40.61 &  \textcolor{red}{\textbf{30.90}} &  \textcolor{red}{\textbf{38.53}} &  \textcolor{red}{\textbf{41.71}}  \\
    & & \cellcolor{mygray-bg}{\textbf{DPL~(Ours)}}   & \cellcolor{mygray-bg}{} & \cellcolor{mygray-bg}{ \textcolor{red}{\textbf{34.87}}} &  \cellcolor{mygray-bg}{\textcolor{red}{\textbf{44.50}}} &  \cellcolor{mygray-bg}{\textcolor{red}{\textbf{47.72}}} & \cellcolor{mygray-bg}{23.54} & \cellcolor{mygray-bg}{30.73} & \cellcolor{mygray-bg}{33.35} \\
    & & \cellcolor{mygray-bg}{\textbf{DPL\textsuperscript{$\dagger$}~(Ours)}}   & \cellcolor{mygray-bg}{} & \cellcolor{mygray-bg}{\textcolor{blue}{31.71}} & \cellcolor{mygray-bg}{\textcolor{blue}{41.56}} & \cellcolor{mygray-bg}{\textcolor{blue}{45.02}} & \cellcolor{mygray-bg}{\textcolor{blue}{24.02}} & \cellcolor{mygray-bg}{\textcolor{blue}{31.26}} & \cellcolor{mygray-bg}{\textcolor{blue}{34.16}} \\
   \cline{2-10}
    & \multirow{5}{*}{5-shot}& RelNet~\cite{sung2018learning} & $_{\textit{CVPR'18}}$  &  18.40 & 25.37 & 28.93 &  15.72 & 20.66 & 23.33  \\
   &  & ProtoNet~\cite{snell2017prototypical}  & $_{\textit{NIPS'17}}$ & 20.37 & 28.36 & 32.77 & 21.75 & 28.22 & 31.56  \\
    & & CoOp~\cite{zhou2022learning}  & $_{\textit{IJCV'22}}$ &  23.18 & 31.56 & 35.10	& 30.11 & 37.47 & 40.54  \\
    & & \cellcolor{mygray-bg}{\textbf{DPL~(Ours)}}  & \cellcolor{mygray-bg}{} &  \cellcolor{mygray-bg}{\textcolor{red}{\textbf{37.48}}} &  \cellcolor{mygray-bg}{\textcolor{red}{\textbf{47.39}}} &  \cellcolor{mygray-bg}{\textcolor{red}{\textbf{50.62}}} & \cellcolor{mygray-bg}{\textcolor{blue}{31.29}} & \cellcolor{mygray-bg}{\textcolor{blue}{39.55}} & \cellcolor{mygray-bg}{\textcolor{blue}{42.72}} \\
    & & \cellcolor{mygray-bg}{\textbf{DPL\textsuperscript{$\dagger$}~(Ours)}}   & \cellcolor{mygray-bg}{} & \cellcolor{mygray-bg}{\textcolor{blue}{35.47}} & \cellcolor{mygray-bg}{\textcolor{blue}{45.86}} & \cellcolor{mygray-bg}{\textcolor{blue}{49.34}} &  \cellcolor{mygray-bg}{\textcolor{red}{\textbf{32.13}}} &  \cellcolor{mygray-bg}{\textcolor{red}{\textbf{40.79}}} &  \cellcolor{mygray-bg}{\textcolor{red}{\textbf{44.24}}}  \\
   \cline{2-10}
    & \multirow{5}{*}{10-shot}& RelNet~\cite{sung2018learning} & $_{\textit{CVPR'18}}$  &  27.63 & 36.72 & 41.05 & 20.36 & 27.92 & 31.83  \\
    & & ProtoNet~\cite{snell2017prototypical}  & $_{\textit{NIPS'17}}$ &  23.32 & 31.41 & 34.98 & 25.15 & 32.72 & 36.30 \\
    & & CoOp~\cite{zhou2022learning} & $_{\textit{IJCV'22}}$  & 23.14 & 29.97 & 33.09 & 29.86 & 37.03 & 40.67  \\
    & & \cellcolor{mygray-bg}{\textbf{DPL~(Ours)}} & \cellcolor{mygray-bg}{}   & \cellcolor{mygray-bg}{\textcolor{red}{\textbf{38.67}}} &  \cellcolor{mygray-bg}{\textcolor{red}{\textbf{49.17}}} &  \cellcolor{mygray-bg}{\textcolor{red}{\textbf{52.43}}} & \cellcolor{mygray-bg}{\textcolor{blue}{32.20}} & \cellcolor{mygray-bg}{\textcolor{blue}{40.33}} & \cellcolor{mygray-bg}{\textcolor{blue}{43.10}} \\
    & & \cellcolor{mygray-bg}{\textbf{DPL\textsuperscript{$\dagger$}~(Ours)}}   & \cellcolor{mygray-bg}{} & \cellcolor{mygray-bg}{\textcolor{blue}{35.66}} & \cellcolor{mygray-bg}{\textcolor{blue}{46.52}} & \cellcolor{mygray-bg}{\textcolor{blue}{50.15}} &  \cellcolor{mygray-bg}{\textcolor{red}{\textbf{32.68}}} &  \cellcolor{mygray-bg}{\textcolor{red}{\textbf{41.26}}} &  \cellcolor{mygray-bg}{\textcolor{red}{\textbf{44.56}}}  \\

\hline\thickhline
\end{tabular}}

\label{tab:VG25-seen-predcls}
\end{table}

\begin{table}[]
    \centering
    \caption{Seen predicate performance (\%) comparison of different methods on VG-25 dataset \wrt SGCls with K-shot. The \textcolor{red}{\textbf{best}} and \textcolor{blue}{second best} results are marked according to formats.}
    \vspace{-1em}
    \scalebox{0.9}{
    \begin{tabular}{ c |c | l l | c c c | c c c}
    \hline\thickhline
    \multirow{2}{*}{D} & \multirow{2}{*}{K-shot} & \multicolumn{2}{c|}{\multirow{2}{*}{Method}} & \multicolumn{3}{c|}{Recall} &  \multicolumn{3}{c}{meanRecall} \\
     \cline{5-10}
      & & & & @20 & @50 & @100  & @20 & @50 & @100  \\
      \hline
      \hline
   \multirow{15}{*}{\begin{sideways}VG-25\end{sideways}}&\multirow{5}{*}{1-shot}& RelNet~\cite{sung2018learning} & $_{\textit{CVPR'18}}$  &  8.79 & 11.24 & 12.26 & 5.91 & 7.94 & 8.90\\
    & & ProtoNet~\cite{snell2017prototypical}  & $_{\textit{NIPS'17}}$ & 9.08 & 11.21 & 12.08 & 6.73 & 8.64 & 9.51 \\
    & & CoOp~\cite{zhou2022learning}  &  $_{\textit{IJCV'22}}$ &  \textcolor{blue}{18.34} & \textcolor{blue}{22.18} & \textcolor{blue}{23.32} &  \textcolor{red}{\textbf{13.53}} &  \textcolor{red}{\textbf{16.18}} &  \textcolor{red}{\textbf{17.13}} \\
    & & \cellcolor{mygray-bg}{\textbf{DPL~(Ours)}}   & \cellcolor{mygray-bg}{} &   \cellcolor{mygray-bg}{\textcolor{red}{\textbf{19.82}}} &  \cellcolor{mygray-bg}{\textcolor{red}{\textbf{23.65}}} & \cellcolor{mygray-bg}{\textcolor{red}{\textbf{24.76}}} & \cellcolor{mygray-bg}{11.28} & \cellcolor{mygray-bg}{14.07} & \cellcolor{mygray-bg}{15.10} \\
    & & \cellcolor{mygray-bg}{\textbf{DPL\textsuperscript{$\dagger$}~(Ours)}}   & \cellcolor{mygray-bg}{} & \cellcolor{mygray-bg}{17.10} & \cellcolor{mygray-bg}{21.01} & \cellcolor{mygray-bg}{22.23} & \cellcolor{mygray-bg}{\textcolor{blue}{11.48}} & \cellcolor{mygray-bg}{\textcolor{blue}{14.30}} & \cellcolor{mygray-bg}{\textcolor{blue}{15.25}} \\
   \cline{2-10}
    & \multirow{5}{*}{5-shot}& RelNet~\cite{sung2018learning} & $_{\textit{CVPR'18}}$  &  9.61 & 11.67 & 12.40 & 11.43 & 13.99 & 14.84 \\
    & & ProtoNet~\cite{snell2017prototypical} & $_{\textit{NIPS'17}}$  &  11.46 & 13.86 & 14.62 &  11.04 & 13.67 & 14.65 \\
    & & CoOp~\cite{zhou2022learning}  & $_{\textit{IJCV'22}}$ & 12.02 & 15.04 & 16.21 & 12.22 & 14.73 & 15.90 \\
    & & \cellcolor{mygray-bg}{\textbf{DPL~(Ours)}}  & \cellcolor{mygray-bg}{} &   \cellcolor{mygray-bg}{\textcolor{red}{\textbf{20.21}}}  &  \cellcolor{mygray-bg}{\textcolor{red}{\textbf{24.09}}} &  \cellcolor{mygray-bg}{\textcolor{red}{\textbf{25.16}}} & \cellcolor{mygray-bg}{\textcolor{blue}{14.29}} & \cellcolor{mygray-bg}{\textcolor{blue}{17.55}} & \cellcolor{mygray-bg}{\textcolor{blue}{18.72}} \\
    & & \cellcolor{mygray-bg}{\textbf{DPL\textsuperscript{$\dagger$}~(Ours)}}  &\cellcolor{mygray-bg}{}  &  \cellcolor{mygray-bg}{\textcolor{blue}{18.21}} & \cellcolor{mygray-bg}{\textcolor{blue}{22.44}} & \cellcolor{mygray-bg}{\textcolor{blue}{23.82}} &   \cellcolor{mygray-bg}{\textcolor{red}{\textbf{14.31}}} &  \cellcolor{mygray-bg}{\textcolor{red}{\textbf{17.57}}} &  \cellcolor{mygray-bg}{\textcolor{red}{\textbf{18.89}}} \\
   \cline{2-10}
   & \multirow{5}{*}{10-shot}& RelNet~\cite{sung2018learning} & $_{\textit{CVPR'18}}$  &  15.84 & 18.81 & 19.70 & 11.67 & 14.37 & 15.31 \\
   & & ProtoNet~\cite{snell2017prototypical}  & $_{\textit{NIPS'17}}$  & 13.39 & 16.82 & 17.97	& 12.23 & 14.96 & 15.93\\
   & & CoOp~\cite{zhou2022learning}  & $_{\textit{IJCV'22}}$ & 10.47 & 13.55 & 14.89 & 10.93 & 13.50 & 14.73 \\
   & & \cellcolor{mygray-bg}{\textbf{DPL~(Ours)}}  & \cellcolor{mygray-bg}{} &  \cellcolor{mygray-bg}{\textcolor{red}{\textbf{20.74}}} &  \cellcolor{mygray-bg}{\textcolor{red}{\textbf{24.77}}} &  \cellcolor{mygray-bg}{\textcolor{red}{\textbf{25.90}}} & \cellcolor{mygray-bg}{\textcolor{blue}{14.62}} & \cellcolor{mygray-bg}{\textcolor{blue}{17.96}} & \cellcolor{mygray-bg}{\textcolor{blue}{19.05}} \\
   & & \cellcolor{mygray-bg}{\textbf{DPL\textsuperscript{$\dagger$}~(Ours)}}  & \cellcolor{mygray-bg}{} & \cellcolor{mygray-bg}{\textcolor{blue}{18.23}} & \cellcolor{mygray-bg}{\textcolor{blue}{22.71}} & \cellcolor{mygray-bg}{\textcolor{blue}{24.23}} &  \cellcolor{mygray-bg}{\textcolor{red}{\textbf{14.78}}} &  \cellcolor{mygray-bg}{\textcolor{red}{\textbf{18.51}}} &  \cellcolor{mygray-bg}{\textcolor{red}{\textbf{19.77}}} \\

\hline\thickhline
\end{tabular}}

\label{tab:VG25-seen-sgcls}
\end{table}

\paragraph{\textbf{Performance Comparison on VG-60 and GQA-50}} We report unseen predicate performance on GQA-50 and VG-60 in Table~\ref{tab:GQA50_predcls} and Table~\ref{tab:VG60_predcls} separately. From tables, we can observe that our method \emph{DPL} has also achieved excellent performance on these two datasets. Compared to VG-25, the unseen predicate categories in VG-60 and GQA-50 have finer-grained semantics, making it more challenging for models to recognize them. Hence we can observe that \emph{DPL} has less advantage compared to the performance on VG-25. Nevertheless, we can still observe that our method shows obvious advantages compared with baseline methods. Specially, \emph{DPL\textsuperscript{$\dagger$}} has achieved the best performance after fine-tuning on support samples.

\paragraph{\textbf{Performance Comparison on SGCls}} We report SGCls performance on VG-25 and VG-60 datasets in Table~\ref{tab:VG-25-unseen-sgcls} and Table~\ref{tab:VG-60-unseen-sgcls}. When performing SGCls prediction, we exploit the predicted object categories to compute attention scores on prototypes. It is more challenging as the object categories for query samples are not given. Hence the accuracy of object prediction also affects the predicate prediction. However from tables, we can see that our method still yields excellent performance.

\subsubsection{\textbf{Seen Predicate Category Performance}} For complete comparison, we also report seen predicate performance in Table~\ref{tab:VG25-seen-predcls} and Table~\ref{tab:VG25-seen-sgcls}. 
Compared to \emph{DPL}, \emph{DPL\textsuperscript{$\dagger$}} has achieved better performance on unseen predicates while dropping a little performance on seen predicate categories. This is caused by the semantic overlapping of predicates in VG dataset. For each subject and object pair, there might be multiple possible predicates but only one will be annotated as ground-truth. The training dataset for \emph{DPL} only consists of seen categories, while the support data set used to fine-tune \emph{DPL\textsuperscript{$\dagger$}} is evenly distributed on both seen and unseen categories. The probability of the model predicting unseen predicates is improved, which inevitably harms the performance of seen predicates. But it still outperforms other baseline methods in most settings. It shows that our method keeps the excellent recognition ability on seen categories while transferring the knowledge to unseen predicate categories.

\subsection{Complexity Analysis}
We also conduct complexity analysis on different methods. We report the average expenses (s) during both the training and inference phases for each image sample, as well as the training parameters (M) in Table~\ref{tab:complexity}. From the table, we can observe that our \emph{DPL} method has a similar complexity to the baseline method in terms of training cost and inference cost, but it has achieved better performance. 
The reason why \emph{CoOp} and \emph{DPL} cost a little more time with fewer parameters is that they need to forward the pretrained text encoder of CLIP. However, by equipping with the knowledge of CLIP, they achieved much better performance on unseen predicates.
While compared to method \emph{CoOp}, \emph{DPL} can be directly used in the following few-shot prediction without fine-tuning on support samples, which is more convenient to the practical application.

\begin{table}[t]
\caption{The Statistics of training cost, test cost and parameters of different methods on 5-shot.}
    \centering
    \vspace{-1em}
    \begin{tabular}{l|c|c| c| c}
    \hline\thickhline
      Model &  Train Cost (s) & Test Cost (s) & Params (M)  & mR@50 \\
      \hline \hline 
      RelNet~\cite{sung2018learning} & 0.0136  & 0.1424 & 320.98 & 20.66 \\
      ProtoNet~\cite{snell2017prototypical} & 0.0129 & 0.1240 & 320.46  & 24.65 \\
      CoOp~\cite{zhou2022learning}   & 0.0193 & 0.1731  & 170.50  & 33.81  \\
      DPL & 0.0167  & 0.1937 & 172.82 & 42.07 \\
      \hline\thickhline
    \end{tabular}
    
    \label{tab:complexity}
\end{table}

\subsection{Study on Decomposed Prototype Learning}
In this part, we mainly verify the effectiveness of different designed components of our proposed prototype generation method. To this end, we conducted experiments under different settings based on PredCls under VG-25, and recorded their unseen predicate performance on Recall@N and mean Recall@N in Table~\ref{tab:fs_ablation}. 
In the experiment, we set $K$ value of $K$-shot as $1$, $5$, and $10$ respectively. \Checkmark in Table~\ref{tab:fs_ablation} means the corresponding component is activated, and \XSolidBrush in Table~\ref{tab:fs_ablation} means the corresponding component is removed.
We mainly compare model performance under the following different designs: 
\begin{itemize}
    \item  \textbf{Learnable Prompts  (\emph{Learn.}):} We utilize learnable prompts to extract semantic prototype representations from the VL model for each predicate, which is opposite to utilizing fixed prompts.
    \item \textbf{Multiple (\emph{Mul.}):} We generate multiple prototype representations for each predicate category instead of only a single prototype representation.
    \item \textbf{Subject-based (\emph{Sub.}):} The subject context information is exploited in the process of prototype generation for each predicate category.
    \item \textbf{Object-based (\emph{Obj.}):} The object context information is exploited in the process of prototype generation for each predicate category.
    \item \textbf{Reweight (\emph{Re-w.}):} The generated prototypes are aggregated by reweight strategy instead of being simply averaged together.
\end{itemize}
We discuss the validity of each component respectively in the following.

\begin{table}[t]
\centering
\caption{Ablation study on decomposed prototypes based on different settings \wrt PredCls with K-shot.}
\renewcommand{\arraystretch}{1.0}
\vspace{-1em}
\scalebox{0.90}{
\begin{tabular}{c | l | c c c c c | c c c | c c c}
\specialrule{0.05em}{0pt}{0pt} 
\hline\thickhline
\multirow{2}{*}{D} &  &\multicolumn{5}{c|}{Settings} & \multicolumn{3}{c|}{Recall} &  \multicolumn{3}{c}{Mean Recall} \\
 & & Learn. & Mul. & Sub. & Obj. & Re-w. &  @20 & @50 & @100  & @20 & @50 & @100  \\
  \hline \hline
  
\multirow{6}{*}{\begin{sideways}1-Shot\end{sideways}} & Single
& \Checkmark  & \XSolidBrush & \XSolidBrush  &  \XSolidBrush & \XSolidBrush & 17.52 & 21.86 & 24.17 & 16.55 & 20.64 & 22.45 \\

&  Fixed  & \XSolidBrush &  \Checkmark & \Checkmark  & \Checkmark  & \Checkmark & 20.05   & 24.10  & 25.81  & 20.85 & 24.92  & 26.70 \\ 
& Average & \Checkmark & \Checkmark & \Checkmark & \Checkmark &  \XSolidBrush &  21.48 & 27.57 & 30.49 & 19.49 & 26.95 & 29.59 \\
& Object & \Checkmark & \Checkmark & \XSolidBrush  & \Checkmark & \Checkmark  & 12.64 & 16.10  & 17.47 &  13.66  &  16.74   &  18.26   \\
& Subject  & \Checkmark & \Checkmark & \Checkmark & \XSolidBrush & \Checkmark &   19.11  &   22.16  &  23.27  &  21.78  &  25.26 &  26.60   \\
& \textbf{DPL} & \Checkmark & \Checkmark & \Checkmark & \Checkmark & \Checkmark  & \textbf{25.42} & \textbf{29.99} & \textbf{31.85} & \textbf{26.15} & \textbf{30.37} & \textbf{32.13} \\

\hline
\multirow{6}{*}{\begin{sideways}5-Shot\end{sideways}} & Single
& \Checkmark  & \XSolidBrush & \XSolidBrush  & \XSolidBrush  & \XSolidBrush  &  29.03	  & 34.60	  & 37.20  & 	29.12	  & 33.81	  & 36.40 \\ 
& Fixed & \XSolidBrush & \Checkmark  &  \Checkmark &  \Checkmark & \Checkmark &  25.79 & 32.48   & 35.25  & 29.57 & 35.68 & 37.86 \\ 
& Average & \Checkmark & \Checkmark & \Checkmark & \Checkmark & \XSolidBrush  &  33.69 & 40.46 & 43.10 & 33.75 & 39.42 & 41.55 \\
&  Object  & \Checkmark & \Checkmark & \XSolidBrush  & \Checkmark & \Checkmark   &  19.70   &  24.94   &  27.15   &  23.08  &  27.53    &  29.60  \\
& Subject  & \Checkmark & \Checkmark & \Checkmark & \XSolidBrush & \Checkmark & 30.64   &  35.45   &  37.54  &  31.63  &  36.04  &  37.82   \\
& \textbf{DPL} & \Checkmark & \Checkmark & \Checkmark & \Checkmark & \Checkmark  & \textbf{37.07} & \textbf{43.30} & \textbf{46.08} & \textbf{36.24} & \textbf{42.07} & \textbf{44.67} \\

\hline
\multirow{6}{*}{\begin{sideways}10-Shot\end{sideways}} & Single
 & \Checkmark  & \XSolidBrush & \XSolidBrush  & \XSolidBrush  & \XSolidBrush & 30.42	  & 36.83  & 	39.95  & 	33.07  & 	39.62  & 42.36 \\
& Fixed & \XSolidBrush & \Checkmark  & \Checkmark  &  \Checkmark & \Checkmark &  23.60 & 30.22   & 32.72  & 27.61 & 33.88 & 35.99 \\ 
& Average & \Checkmark & \Checkmark & \Checkmark & \Checkmark & \XSolidBrush &  37.36 & 44.25 & 47.29 & 34.17 & 40.36 & 43.03 \\
&  Object  & \Checkmark & \Checkmark &  \XSolidBrush & \Checkmark & \Checkmark  &  25.19   &  30.05   &  32.05  &  26.81  &  31.07   & 32.84   \\
&  Subject  & \Checkmark & \Checkmark & \Checkmark & \XSolidBrush & \Checkmark &  30.59   &  35.98   &  38.09  &  32.63  &   38.09  &  40.05   \\
& \textbf{DPL} & \Checkmark & \Checkmark & \Checkmark & \Checkmark & \Checkmark  &  \textbf{40.31} & \textbf{47.02} & \textbf{49.53} & 	\textbf{39.35} & \textbf{45.66} & \textbf{48.15} \\

\hline\thickhline
\end{tabular}}

\label{tab:fs_ablation}
\end{table}

\subsubsection{\textbf{Multiple prototypes \vs Single prototype}}
To demonstrate the effectiveness of the decomposed learning of multiple prototypes for each predicate category, we conducted the experiment by replacing multiple prototypes with only a single prototype (\eg, \emph{CoOp}). The performance is shown in the 1st line of each K-shot setting in Table~\ref{tab:fs_ablation}.

We can see that the performance of multiple prototype representation (\cf last line ) is significantly better than that of single prototype representation. This is because the multiple prototype representation method introduces information about the subject-object pairs from support samples, which can help the model more precisely learn diverse semantic knowledge of predicate categories and different subject-object composition patterns.

\subsubsection{\textbf{Learnable prompts \vs Fixed prompts}}
In order to demonstrate the superiority of our devised learnable prompts, we conducted the experiment by replacing learnable prompts with fixed prompts, such as ``\texttt{This is a photo of [cls]}". The performance is reported in the 2nd line of each K-shot setting in Table~\ref{tab:fs_ablation}.

We can see that our learnable prompts (\cf last line) have greatly improved the model performance compared to the fixed prompt.
Well, directly extracting semantic knowledge using fixed prompts will harm the model performance on the unseen categories, which illustrates the difficulty of the target to model the semantic prototype of predicates. It is hard to directly use artificial prompts to extract knowledge of unseen predicate categories. In contrast, the learnable prompts we designed can more effectively extract the knowledge of the VL model and bring benefits to the model performance.
Compared to the performance of single prototype on meanRecall, fixed prompt shows obvious advantages on 1-shot and 5-shot, which also verifies the effectiveness of our decomposed prototype module. However, the performance of the fixed prompt drops significantly under the 10-shot setting, which is mainly because the fixed prompt with the subsequent reweight aggregation strategy cannot effectively handle the difference in complex support samples.

\subsubsection{\textbf{Average aggregation \vs Reweight aggregation}}
For multiple prototype representations of each predicate category, we reweight them according to the input query samples with adaptive assigned weights to generate more reliable category representations.
In order to verify the effectiveness of the reweight aggregation strategy, we used two different aggregation strategies to conduct experiments and recorded the experimental results in the 3rd line of each K-shot setting of Table~\ref{tab:fs_ablation}.

From the table, we can observe that the reweight aggregation strategy performs significantly better than the average aggregation strategy on both Recall@N and meanRecall@N. This is because the reweight strategy is able to cope with the semantic and visual differences within predicate categories, thereby generating more reliable category representation for each query sample. Compared with single prototype, we can see that using multiple prototypes to represent the same predicate category even with average strategy can still achieve better performance.

\subsubsection{\textbf{Subject-based \vs Object-based Prototypes}}
To verify the validity of using subject and object context when constructing prototypes, we conducted experiments by removing them respectively. The performance is shown in the 4th and 5th lines of each K-shot setting in Table~\ref{tab:fs_ablation}.

We can see that removing either subject or object information when constructing prototypes will harm model performance, which verifies the effectiveness of the subject and object context when constructing prototypes. 
Besides, compared to the performance of single prototype (\cf 1st line), multiple prototypes constructed with only object or subject information perform worse in some cases. Such observation shows that the semantics of each predicate depend on both subject and object simultaneously and only considering either one is not able to improve performance.  
Compared to the 4th line and 5th line results, we can also see that subject information may provide more useful clues for making accurate predictions on VG-25 dataset.

\begin{figure*}[t]
  \centering
  \includegraphics[width=0.97\linewidth]{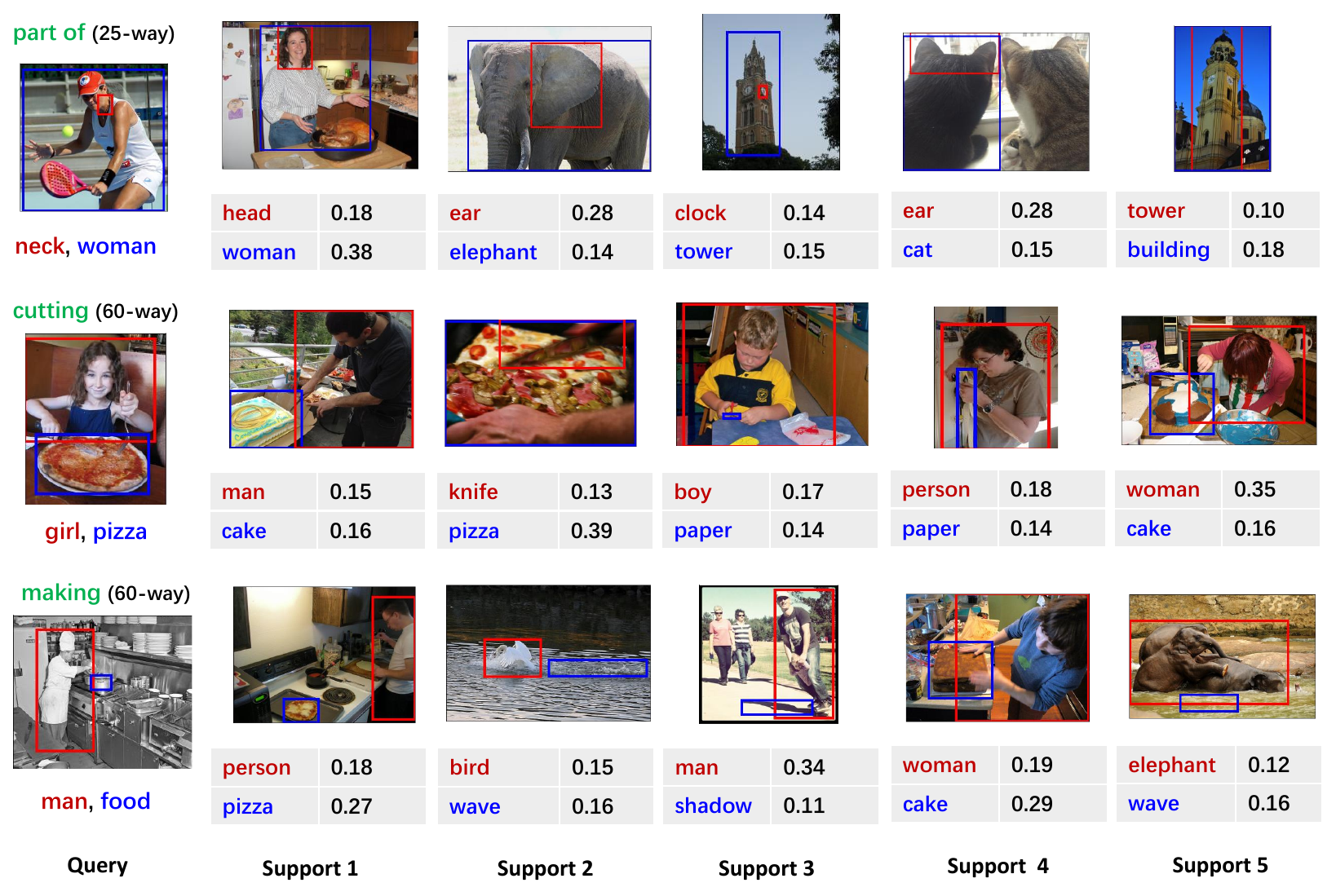}
  \caption{Visualization of the assigned weights to the prototypes based on $5$-shot of support samples. The first column displays the query triplets and the target predicates (highlighted in \textcolor{ForestGreen}{green}). The other columns display the support triplets and their assigned weights by DPL. The subjects and objects are drawn in \textcolor{red}{red} boxes and \textcolor{blue}{blue} boxes respectively.}
  \label{fig:visualize}
\end{figure*}

\subsection{Visualization}
To show how the prototype weighted aggregation strategy works, we randomly selected some query samples from the 25-way and 60-way test sets respectively, and presented their weight assignment details in Fig.~\ref{fig:visualize}. We visualized the assigned weights by the DPL model to each prototype representation generated from different support samples (including subjects and objects) during the prediction process. Among them, the first column shows the query sample and the target predicate category required to be recognized (the text highlighted with \textbf{\textcolor{ForestGreen}{green}} in the picture). The other columns show the 5-shot support triplet samples of the category and the weights assigned by the model to the prototype representations generated from different support samples. The subject and object are circled in the figure with \textbf{\textcolor{myred}{red}} and \textbf{\textcolor{blue}{blue}} boxes respectively.
From Fig.~\ref{fig:visualize}, we can observe the following information:

The visual appearance of each support sample of the same predicate category varies greatly under different subject-object pair combinations. For example, the visual appearance of the triplet "head $\rightarrow$ part of $\rightarrow$ woman" is completely different from the visual appearance of the triplet "clock $\rightarrow$ part of $\rightarrow$ tower", hence the model needs to generate a truly reliable prototype representation for the query sample while making predictions.

For each query sample, our model tends to assign higher weights to subject-object prototypes with closer semantics, thereby generating more reasonable category representations for them.
Take the query sample of "cutting" in 60-way as an example. For the subject prototype, the model assigns a higher weight to the support sample $5$, while for the object prototype, the model assigns a higher weight to the support sample $2$. In this way, our method can better handle the impact of predicate intra-class variance and generate more reliable predicate category prototypes for each sample, thus making more accurate predictions. Here we only consider the semantic connections between subjects or objects. In future work, we will consider incorporating more useful information into the reweighting process to generate more reliable prototypes.

\section{Conclusion}
In this paper, we focus on a promising and important research task: Few-shot Scene Graph Generation (FSSGG), which requires SGG models to be able to recognize unseen predicate categories quickly with only a few samples. Different from existing FSL tasks, the multiple semantics and great variance of the visual appearance of predicate categories make FSSGG non-trivial. To this end, we devised a novel Decomposed Prototype Learning (DPL). Specifically, we introduced learnable prompts to model diverse semantics of predicates with the help of pretrained VL models by decomposing each predicate into multiple prototypes. Afterward, we adaptively assigned weights to these prototypes by considering the subject-object compositions to generate query-adaptive prototype representation.
To verify the effectiveness of our methods, we conduct extensive experiments on multiple datasets under different few-shot settings.
Our decomposed prototype learning method can be effectively applied to other FSL researches that need to address high intra-class variation in scene understanding tasks.
One limitation of our work is that it does not consider the redundancy between prototype representations generated from similar object categories, which may lead to confusion when applying attention scores for each prototype. To eliminate this redundancy, we may consider clustering similar prototypes by using multi-modal features of object categories or leveraging the knowledge of Large Language Models. We leave this as our future work.

\section*{Acknowledgments}
This work was supported by the National Key Research \& Development Project of China (2021ZD011 \\ 0700), the National Natural Science Foundation of China (U19B2043, 61976185),  and the Fundamental Research Funds for the Central Universities (226-2022-00051). Long Chen was supported by HKUST Special Support for Young Faculty (F0927) and HKUST Sports Science and Technology Research Grant (SSTRG24EG04)



\bibliographystyle{ACM-Reference-Format}
\bibliography{references}

\end{document}